 \documentclass[final,5p,times,twocolumn]{elsarticle}

\usepackage{amssymb}

\usepackage{algorithm}
\usepackage{algpseudocode}
\usepackage{graphicx}
\usepackage{lscape}
\usepackage{framed}
\usepackage{url}
\usepackage{color, colortbl}
\usepackage[table]{xcolor}
  \usepackage{soul, color}
  \sethlcolor{green}
\usepackage{comment}
 \usepackage[neveradjust]{paralist}
 \usepackage{amsmath}
\usepackage{amssymb}
\usepackage{amsthm}
\usepackage{stmaryrd}

\usepackage{listings}

 \usepackage{multicol}
 
\usepackage{tabularx}

\usepackage{enumitem}
\usepackage{mathtools}
\usepackage{cancel}
\usepackage{centernot}

\usepackage{xspace}
\usepackage{hyphenat}
\usepackage{array}

\usepackage{graphicx}
\usepackage{multirow}

\usepackage{varwidth}
\usepackage{xcolor}
\usepackage{lipsum}

\usepackage{subcaption}

\usepackage{alltt}

\newcommand{\typedliteral}{\textasciicircum\textasciicircum}

\algnewcommand{\Input}[1]{\item[\textbf{Input:}] \parbox[t]{\linewidth-1cm}{#1}\vspace{3pt}}
\algnewcommand{\And}{\textbf{and}\xspace}
\algnewcommand{\Or}{\textbf{or}\xspace}
\algnewcommand{\Not}{\textbf{not} }
\algnewcommand{\To}{\textbf{to}\xspace}

\let\oldtextproc\textproc
\renewcommand{\textproc}[1]{\nohyphens{\oldtextproc{#1}}}

\usepackage{soul, color}
\sethlcolor{green}

\definecolor{jp}{RGB}{192, 36, 19}

\journal{TBA}

\begin{document}

\begin{frontmatter}



\title{Competency Questions and SPARQL-OWL Queries Dataset and Analysis}


   \author[put]{Dawid Wi\'sniewski\corref{cor1}}
   \ead{dawid.wisniewski@cs.put.poznan.pl}
         \author[put,camil]{Jedrzej Potoniec}
         \ead{jedrzej.potoniec@cs.put.poznan.pl}
    \author[put,camil]{Agnieszka \L{}awrynowicz}
    \ead{agnieszka.lawrynowicz@cs.put.poznan.pl}
             \author[cmk]{C. Maria Keet}
             \ead{mkeet@cs.uct.ac.za}
         
     \address[put]{Faculty of Computing, Poznan University of Technology, ul. Piotrowo 2, 60-965 Poznan, Poland}
         \address[camil]{Center for Artificial Intelligence and Machine Learning, Poznan University of Technology, ul. Piotrowo 3, 60-965 Poznan, Poland}
    \address[cmk]{Department of Computer Science, University of Cape Town, Private Bag X3
    	Rondebosch
    	7701
    	South Africa }
    
    \cortext[cor1]{Corresponding author}
 
%

\begin{abstract}
Competency Questions (CQs) are natural language questions outlining
and constraining 
 the scope of knowledge represented by an ontology. 
Despite that CQs are a part of several ontology engineering methodologies, we have 
observed that 
the actual publication of CQs for the available ontologies is 
very limited
and even scarcer is the publication of their respective formalisations in terms of, e.g., SPARQL queries. 
This paper aims to contribute to addressing the engineering shortcomings of using CQs in ontology development, to facilitate wider use of CQs.
In order to understand the relation between CQs and the queries over the ontology to test the CQs on an ontology, we gather, analyse, and publicly release a set of 234 CQs and their translations to SPARQL-OWL for several ontologies in different domains developed by different groups.
We analysed the CQs in two principal ways. The first stage focused on a linguistic analysis of the natural language text itself, i.e., a lexico-syntactic analysis without any presuppositions of ontology elements, and a subsequent step of semantic analysis in order to find patterns.
This increased diversity of CQ sources resulted in a 5-fold increase of hitherto published patterns, to 106 distinct CQ patterns, which have a limited subset of few patterns shared across the CQ sets from the different ontologies. 
Next, we analysed the relation between the found CQ patterns and the 46 SPARQL-OWL query signatures, which revealed that one CQ pattern may be realised by more than one SPARQL-OWL query signature, and vice versa.
We hope that our work will 
contribute to 
establishing common practices, templates, automation, and user tools that will support
CQ formulation, formalisation, execution, and general management. 

\end{abstract}


\end{frontmatter}

\section{Introduction}
\label{sec:intro}

Within the field of ontology engineering,  
\emph{Competency Questions (CQs)}~\cite{DBLP:journals/ker/UscholdG96} are natural language questions outlining the scope of knowledge represented by an ontology. 
They represent functional requirements in the sense that the developed ontology or an ontology-based information system should be able to answer them; hence contain all the relevant knowledge.
For example, a CQ may be \emph{What are the implementations of C4.5 algorithm?}, indicating that the ontology needs to contain classes, such as \textsf{Algorithm} and \textsf{C4.5} as subclass of \textsf{Algorithm}, and something about implementations such that the answer to the CQ will be non-empty. 

CQs are 
a part of several ontology engineering methodologies, yet the actual publication of CQs for the available ontologies is rather scarce. Even more scarce is the publication of the CQs' respective formalisation in terms of, e.g., SPARQL queries. This suggests CQs are not used widely as intended. 
We hypothezise that it may be due to the lack of common practices, templates, automation, and user tools that would support CQ formulation, formalisation, execution, and general management; or: it is still a fully manual process. For instance, even if one has specified CQs, there is no automatic way to translate it to, say, a SPARQL-OWL~\cite{DBLP:conf/esws/KolliaGH11} query (for validation and verification), and not even a systematic manual way either.

There have been few attempts to analyse CQs.  
Ren et al.~\cite{DBLP:conf/esws/RenPMPDS14} analysed CQs and their patterns to determine CQ archetypes, as tried \cite{Bezerra14}. Those patterns have a limited coverage, however, for they are based on CQ sets of at most two ontologies (Pizza and Software), which thus may contain domain bias, CQ author bias, and `prejudiced' patterns as the Pizza CQs were created after the ontology. 
As simple example of the latter issue, one could create a CQ \textit{Which pizza has hot as spiciness?} that neatly fits with Pizza's \textsf{hasSpiciness} data property, or a more natural phrase \textit{Which pizzas are hot?} that is fully agnostic of how it is represented in the ontology, be it with a data property, object property, or a class. More generally, it suggests that Ren et al.'s CQ patterns, formulated alike ``Which [CE1] [OPE] [CE2]?'', may not be appropriate as CQ pattern, as it presupposes which kind of element it would be in an ontology. 
The manual process and `free form' formulation of CQs by domain experts also runs onto problems that some turn out not translatable into a test over the ontology for various reasons. For instance,
the CQ \textit{How can I get problems [with X] fixed?} of the Software Ontology cannot be answered by a declarative specification that the ontology is, or 
take the CQ for the DMOP ontology \cite{DBLP:journals/ws/KeetLdKNPSH15}: \emph{Given a data mining task/data set, which of the valid or applicable workflows/algorithms will yield optimal results (or at least better results than the others)?}: 
assuming that the question may deal with an arbitrary (not pre-defined upfront) dataset, 
this CQ may only be answered via performing data mining experiments and not by the ontology itself. 
Therefore, without a clear guidelines of what kind of CQs may be meaningfully expressed and used as requirement specification for an ontology's content, their uptake and usage likely will remain limited. 
This paper aims to contribute to addressing the engineering shortcomings of using CQs in ontology development.

To clear up the CQ muddle and trying to understand the relation between CQs and the queries over the ontology to test the CQs on an ontology, we gather, analyse, and publicly release a larger set of CQs and their translations to SPARQL-OWL for several ontologies in different domains developed by different groups. 
%
%
For the analysis in particular, we seek to address the following research questions:
%
%
\begin{description}
\item[RQ1:] Increasing the scope in domains and ontologies, are there more CQ patterns than those identified in the state-of-art papers?
\item[RQ2:] Are the linguistic patterns specific to constructs of OWL?
\item[RQ3:] Are there recurring query signatures at SPARQL-OWL level? 
\item[RQ4:] How do the linguistic patterns of CQs link to SPARQL-OWL query signatures?
\end{description}

A total of 234 CQs for 5 ontologies have been collected and translated into SPARQL-OWL queries, and made available as a data resource. We analysed them in two principal ways. The first stage focused on a linguistic analysis of the natural language text itself, i.e., a lexico-syntactic analysis without any presuppositions of ontology elements, and a subsequent step of semantic analysis. This revealed 17 CQ patterns at the natural language layer. While a few patterns occur in multiple CQ sets, there are also patterns unique to a CQ set, supporting the expectation that a broad sampling is required to obtain a more representative set of patterns. 
The second phase consists of designing SPARQL-OWL queries for all CQs, where possible, and examining the signature of the queries. We found 46 query signatures resulting from the collected 131 SPARQL-OWL queries. 
The third step consists of the analysis of the relation between the CQ patterns and the SPARQL-OWL query signatures. This is, as hypothesised, a $m$:$n$ relation, or: one CQ pattern may be realised by more than one SPARQL-OWL query and there may be more than one CQ pattern for a SPARQL-OWL query signature.
%

The remainder of the paper is structured as follows. We first discuss related works on CQs and CQ patterns in Section~\ref{sec:relworks}. Section~\ref{sec:cq:lang} is devoted to the linguistic analysis of CQs and Section~\ref{sec:cq:sparql} to the generation and analysis of the SPARQL-OWL queries. We discuss and return to the research questions in Section~\ref{sec:disc} and conclude in Section~\ref{sec:concl}. The data is available from a Git repository at \url{https://github.com/CQ2SPARQLOWL/Dataset}.


\section{Related works}
\label{sec:relworks}

We first discuss related work on CQ-driven ontology authoring, to proceed to a detailed analysis of the current most comprehensive categorisation of CQs, and closing with a discussion on CQs on modelling styles in an ontology. 

\subsection{Competency Question-driven Ontology Authoring}

Competency questions (CQs) specify what knowledge has to be entailed in the ontology and thus can be seen as a set of requirements on the content as well as a way of scoping and delimiting the subject domain that has to be represented in the ontology. They were introduced in  
\cite{DBLP:journals/ker/UscholdG96} and are included in several ontology engineering methodologies~\cite{DBLP:journals/ao/Suarez-Figueroa15,fernandes2011using}. There is, however, a scant publication record of CQs for ontologies. Notable comprehensive CQ sets are those for the Software Ontology 
\cite{Malone14} and Dem@Care \cite{DEMCARE}, but, mostly, ontology papers typically list only a few illustrative CQs, if at all (a few can be found in, e.g., \cite{DBLP:journals/ws/KeetLdKNPSH15} for DMOP and the OWL-formalised SAREF4Health model \cite{Moreira18}). A lack of a representative set of CQs for individual ontologies as well as for ontologies across domains hampers methodological and tooling support for CQs, which, in turn, may hamper specification of CQs in the ontology development process. 

Ren et al. \cite{DBLP:conf/esws/RenPMPDS14}  analysed the structure of CQs and propose a set of 19 ``archetypes'' of CQs that have a flavour of CQ templates that  a domain expert could fill in; e.g., the ``Which [CE1] [OPE] [CE2]?'' archetype where one could fill in the `slots' with, say, ``Which [animal] [eats] [fruit]?'' for an ontology about pets. Similarly, Bezerra et al. \cite{Bezerra14} propose 14 patterns to function as a Controlled Natural Language (CNL) by means of templates for CQs; e.g. ``Does $<$class$>$ + $<$property$>$ $<$class$>$?'' that could be filled in with vocabulary from the ontology, alike {\em Does  [giraffe] [eat] [fruit]?}. We shall discuss in some detail the categorisation of Ren et al.'s archetypes. Generally, from a CQ usability viewpoint, their work is paper-based and a knowledge engineer still has to formulate manually the SPARQL or SPARQL-OWL queries out of the CQs. CQChecker is a tool that first checks whether the CQ is over classes and relations (TBox), over instances (ABox) or is a true/false decision problem (with as example ``Is pizza a food?''), and sends the query in the form of a description logic query to Pellet or executes a SPARQL query, respectively. Technical details on the query formulation from the CNL are not described, however, nor is the basis on which the templates were devised (other than having them ``observed'').  

Related to the queries that intuitively goes hand-in-hand with CQs, is that evaluating a CQ over an ontology may be seen as a {\em test} on the ontology to check whether the knowledge is entailed or what its contents are. The notion of `testing' an ontology in one way or another is not new \cite{Vrandecic06}, and recent proposals include  testing an axiom  \cite{KL16,Warrender15}, the quality of the data associated with it \cite{Kontokostas14}, and manually specified ``test expressions'' based on a manual categorisation of 248 CQs into 10 types. 
Checking whether a CQ---assumed to have been formulated by domain experts---holds in the ontology amounts to {\em validation}, which is different from such efforts that aim to solve other problems, such as verification and authoring (therefore, those techniques also do not apply here). The focus here is to understand CQs better toward better engineering support for them.

\subsection{Analysis of existing categorisation of CQ types}
\label{sec:RenEtAl}

Ren et al.~\cite{DBLP:conf/esws/RenPMPDS14} analysed CQs and their patterns to derive CQ archetypes and mappings from each to a set of the ontology authoring tests. It grouped types of questions in the following way:
%
%
%
\begin{enumerate}
\item QuestionType (QT) - is the question of selection type, binary yes/no or counting - expecting a number to be calculated?
\item ElementVisibility (EV) - are class expressions and property expressions expressed in a CQ in an explicit or implicit manner?
\item QuestionPolarity (QP) - is the question expressed in a positive or negative manner?
\item PredicateArity (PA) - what is the number of arguments of the main predicate?
\item RelationType (RT) - is the relation a CQ contains a datatype or object property?
\item Modifier (M) - Does CQ contain a quantity or numeric modifier?
\item Domain-independent element (DE) - does the CQ contain element that can occur in different domains? Like, is the question about place or time? 
\end{enumerate}
However, only QT, QP, and M can be properly measured independent of the actual ontology. The others are {\em dependent on the ontology}; hence, it is not possible to categorise CQs fully without an existing ontology. Let us illustrate the issues. 

The relation type (RT) can only be chosen when an ontology engineer has decided how to represent the knowledge. For instance, the CQ set for the Software Ontology includes, among others,  ``What is the homepage of Weka?''. The relation \emph{is the homepage of} can be extracted from the CQ and then mapped to the \textsf{hasWebsiteHomepage} data property in the  ontology, with \texttt{rdfs:Literal} as its range. Yet, the ontology also contains the class \textsf{URL}, whose instances could be valid arguments of the relation if the ontology engineer decided so with an object property. Thus, on the CQ level (or stage in the development process) only, it is not possible to obtain the value of the RT feature. 

Second, consider PredicateArity (PA), which is hard to define on the level of CQ itself. Ren et al.~\cite{DBLP:conf/esws/RenPMPDS14} define a unary predicate as containing ``a single set of  entities/values and their instances'' with the example of a unary CQ: ``Is it thin or thick bread?''. However, the knowledge easily can be represented in different ways; e.g.: 
\begin{itemize}
\item \textsf{thick(bread)} - being an unary predicate.
\item \textsf{hasThicknessLevel(bread, thick)} - being a binary predicate.
\end{itemize}
While one can bring afore arguments why one way of representing the knowledge would be better ontologically than another, that fact is that the language permits both options. An actual example can be found in Software Ontology for the given CQ ``Does Weka provide XML editing?'':
\begin{itemize}
\item \textsf{provide(Weka, XML editing)} - with \textsf{provide} being a binary predicate joining a piece of software with given process.
\item \textsf{provideXMLEditing(Weka)} - with \textsf{provideXMLEditing} functioning as a unary predicate defining the \textsf{provideXMLEditing} as a quality of \textsf{Weka}.
\end{itemize}

Thus, also in this case, the way that a CQ is expected to be interpreted (and classified into one of those 7 types) depends on how the knowledge is modelled in the ontology.

The other two problematic ones, element visibility (EV) and domain-independent element (DE), require a slightly deeper analysis. For EV, although there are clear cases where it does work out, even Ren et al.'s example is problematic. 
The CQ ``What are the export options for this software?'' is purported to have the explicit classes \textsf{export option} and \emph{software} and implicit object property \textsf{hasExportOption}. Yet, the \textsf{are the export option of} fragment of CQ or its normalised singular form \textsf{is the export option for} can also be an object property---as its inverse, in fact---which is explicitly mentioned in the CQ and can be expressed even with the same words in an ontology. As in the description of the PA issue, there is an actual CQ defined for the Software Ontology that exhibits this issue: the CQ ``Is Matlab FOSS?'' can be interpreted in different ways:
\begin{itemize}
\item Assume that the ontology contains a class \textsf{Matlab} and a relation named \textsf{isFoss} (with \texttt{rdfs:range} \texttt{xsd:boolean} defined). Both `Matlab' and `is FOSS'  occur explicitly in the CQ, thus the feature value should be set to explicit.
\item Assume that the ontology contains a class \textsf{FOSS} being subtype of \textsf{LicenseClause}, \textsf{Matlab} of type \textsf{Software} and property \textsf{hasLicenceClause} with \textsf{Software} in its domain and \textsf{LicenseClause} in its range. In that case we should mark our CQ as having implicit relation, because the name of it does not appear explicitly in the CQ. 
\end{itemize}
%
Lastly, the `domain independence' in DE is difficult to operationalise. 
For instance, causality is a domain-independent notion, as is parthood. Ontologists have not agreed on which elements are truly domain-independent, so then the value of this feature for a given CQ would depend on the modeller's viewpoint or the foundational ontology the modeller may wish to use for the domain ontology.

Thus, overall, while the idea of classifying CQs to structure what is going on at that stage of ontology development may be useful, the currently best (and only) list available faces several issues, the principal one being that knowledge can be represented differently in an ontology, and sometimes even in the same ontology. This we shall discuss next.


\subsection{CQs and Modelling Styles}



The aforementioned issues with RT, PA, and EV are due to differences in representing knowledge in the ontology even after counting for different names or labels (e.g., dashes vs camel case, names vs identifiers with labels). The axiom type(s) used are modelling choices that, when applied repeatedly and consistently throughout the ontology, becomes a `style' used in an ontology. For instance, whether to use a branch for qualities that are then classes (called `value partition' in the popular Pizza ontology) or to use data properties, and whether to have a branch in the class hierarchy for processes (perdurants/occurrents) to which one relates with two new relations the participating objects (endurants/continuants) from a separate branch, or add an object property and relate the two classes through that property. A typical example of the latter is \textsf{marriage} as a class---typical for an ontology inspired by a foundational ontology---or \textsf{married to} as an object property, and of the former a quality like \textsf{colour}. Five such common alternate modelling patterns were presented in \cite{FK17}, which is likely still an incomplete list. Ideally, a decision for one or the other pattern is made for all such cases in an ontology in the same way. 

CQs are, in theory at least, independent of the modelling style chosen, because they are supposed to be defined upfront according to the extant methodologies that include such a step. Yet because of the different modelling styles, it then implies that a CQ may need to map to different queries, for the queries in, say, SPARQL, {\em are} tailored to the ontology's content. Therefore, a CQ does not {\em a priori} indicate the axiom type with corresponding single query pattern, and, hence there cannot be usable ``CQ Archetypes'' of the form as presented in Ren et al. and Bezerra et al., alike a ``How much does [CE] [DP]?'', because at the CQ level there is {\em natural language}, but  no modelling decisions on OWL constructs and modelling style. 
What CQ patterns do look like and how they map to queries is a question that remains to be answered.

%

\section{Analysis of Competency Questions}
\label{sec:cq:lang}
The aim of the analysis of the CQs is to examine whether there are some popular linguistic structures that can be reused to specify requirements for, and validate, new and existing ontologies.
This section describes the collection of the materials, the methods, and subsequently the results of the CQ analysis.

\subsection{Materials and Methods}

We describe and motivate the materials first and then proceed to the methods and motivations thereof.

\subsubsection{Materials}

There are multiple ontologies available over internet with competency questions provided, but since the focus of our research is on SPARQL-OWL queries, we selected only those ontologies with CQs stated against ontology schema (T-Box). As a result we selected 5 ontologies with 234 competency questions in total. Table \ref{table:cq_benchmark} summarizes our dataset size and source of each ontology.


\begin{table}[h]
\begin{center}
\caption{Competency questions dataset summary}
\label{table:cq_benchmark}
\centering
 \begin{tabular}{p{.7\linewidth}>{\raggedleft\arraybackslash}p{.19\linewidth}} 
 \hline
 Name (short name) & CQ count \\ [0.5ex] 
 \hline\hline
 Software ontology (SWO)~\cite{SWO} & 88 \\ 
 \hline
 Stuff ontology (Stuff)~\cite{10.1007/978-3-319-13704-9_17} & 11 \\
 \hline
 African Wildlife ontology (AWO)~\cite{Keet18oebook} & 14 \\
 \hline
 Dementia Ambient Care ontology (Dem@Care)~\cite{DEMCARE} & 107 \\
 \hline
  Ontology of Datatypes (OntoDT)~\cite{Panov:2016:GOD:2869973.2870281} & 14 \\
  \hline
\end{tabular}
\end{center}
\end{table}

The Software Ontology (SWO) \cite{SWO} is included because its set of CQs is of substantial size and it was part of Ren et al.'s set of analysed CQs. The CQ sets of Dem@Care \cite{DEMCARE} and OntoDT \cite{Panov:2016:GOD:2869973.2870281} were included because they were available. CQs for the Stuff \cite{10.1007/978-3-319-13704-9_17} and African Wildlife (AWO) \cite{Keet18oebook} ontologies were added to the set, because the ontologies were developed by one of the authors (therewith facilitating in-depth domain analysis, if needed), they cover other topics, and are of a different `type' (a tutorial ontology (AWO) and a core ontology (Stuff)), thus contributing to maximising diversity in source selection.

\subsubsection{Methods}

The methods involve various NLP tasks and analysis of the outputs.

\paragraph{Chunks and pattern candidates}
In order to identify regularities among collected questions, we analyzed the linguistic structure of every CQ. Because the dataset does not contain any pair of questions consisting of identical sequences of words, we proposed a pattern detection procedure to identify more general groups that will be the focus of our analysis.

We observed that some CQs share the same structure, but use different vocabulary. For example:
\begin{enumerate}
\item CQ from AWO: \emph{Which plants eat animals?}
\item CQ from SWO: \emph{Which software tool created this software?}
\end{enumerate}
Both of the questions share the same structure: \emph{Which [subject] [predicate] [object]?}, but different subject, predicate and object are used.
Because different ontologies describe various domains, they are likely to contain different vocabularies. Thus CQs stated against those ontologies will contain different vocabularies.


To address the issue, we generate ontology vocabulary agnostic patterns. The pattern candidate generated from a  CQ is a CQ with vocabulary being likely ontology specific replaced with artificial identifiers. Having such patterns we can observe regularities shared among multiple CQs and different ontologies.

We propose two kinds of artificial identifiers:
\begin{itemize}
\item Entity Chunk (EC): fragment of text describing an object (entity) that is likely to be represented in the ontology.
\item Predicate Chunk (PC): fragment of text being a simple predicate  that represent relations between entities that are likely to be represented in the ontology.
\end{itemize}

In order to create a SPARQL-OWL query for a given pair of CQ and ontology, ECs and PCs from a CQ should be matched with appropriate vocabulary from an ontology. We observed that in some cases the ontology contains synonyms that should be matched. As an example, swo14: \emph{Which software tool created [this data]?} - is stated against SWO ontology which vocabulary does not contain any entry named \textsf{created}, but there is \textsf{has specified data output} property that in the given context can be chosen instead.

Thus,  on the level of linguistic analysis of CQs, it cannot be guaranteed that some chunk will be represented in an ontology, and sometimes different phrase will be appropriate. That is why we used \emph{is likely to be represented} to describe chunks.

The chunk identification and replacing procedure is defined as follows:
\begin{enumerate}
\item Perform a tokenization, part-of-speech (POS) tagging and dependency tree parsing on a CQ, so each word obtains a part of speech and dependency tags. For that purpose, SpaCy software\footnote{\url{https://spacy.io/}} is used.
\item Use rules\footnote{\url{https://tinyurl.com/y89hqzol}} to identify longest matched sequences of words with expected POS-tags and/or structure obtained from dependency tree.
\item Use dependency tree parsing in order to identify auxiliary verbs for PCs. Because PC can be discontinuous, having an auxiliary verb separated from the sequence of words detected as PC -- check if any word from the PC is connected with an auxiliary word with an 'aux' relation label. If so, enrich PC with that auxiliary word.
\item Replace all detected ECs with EC text, followed by a numeric identifier, unique for every EC chunk in that CQ.
\item Replace all detected PCs with PC text followed by a numeric identifier, unique for every PC chunk in that CQ.
\item Validate the quality of given substitutions to fix incorrect tags if occur due to imperfect POS-tagger and dependency tree parser usage.
\end{enumerate}

As a result, questions like awo\_6: \emph{Which plants eat animals?} become: \emph{Which EC1 PC1 EC2?}, since \emph{plants} and \emph{animals} are identified as ECs by rules, and \emph{eat} is identified as a PC.

For CQs containing auxiliary verbs, like awo\_4: \emph{Does a lion eat plants or plant parts?} the process transforms it into: \emph{PC1 EC1 PC1 EC2 or EC3?}, because \emph{a lion}, \emph{plants} and \emph{plant parts} are identified as ECs, and \emph{eat} with an auxiliary verb \emph{does} is interpreted as single PC.

The remainder of this subsection focuses on analysis of patterns created using chunk replacement.

\paragraph{Pattern candidate filtering}
\label{filtering_sect}


It is important to mention, that apart of Dem@Care all other collected ontologies have some number of CQs containing placeholders (In fact, for SWO and OntoDT all CQs contain placeholders).
A placeholder is a fragment of text, that should be filled with some class or instance from the ontology. It is introduced to indicate that the CQ makes sense for multiple placeholder fillings and each possible filling may be used to produce a CQ.
For example, for CQ swo06: \emph{Does [this software] provide XML editing?} \emph{[this software]} is a placeholder, which can be replaced with some actual subclasses of software, so multiple CQs can be obtained, like:
\begin{itemize}
\item \emph{Does Weka provide XML editing?}
\item \emph{Does Ringo provide XML editing?}
\item \emph{Does Microsoft Windows provide XML editing?}
\end{itemize}

We call the form of CQ with placeholders filled with actual data from an ontology a materialized form, while CQ with placeholder without filling -- dematerialized form.

All possible fillings of placeholders in our dataset are interpreted by our rules as EC, so if a CQ contains a placeholder, we assume that it can be replaced with EC at pattern candidate construction stage.

The pattern, by definition should be something that is regularly repeated arrangement. Thus a pattern candidate in order to be accepted as a pattern should be observed more than once in a dataset. Because some CQs are materialized and some not, we defined different procedures of accepting pattern candidates as patterns.

\begin{itemize}
\item for dematerialized CQ: each dematerialized CQ has a placeholder that can be potentially filled in multiple ways. Thus single dematerialized CQ can produce multiple materialized CQs. Because of that, every pattern candidate produced out of dematerialized CQ is interpreted as a pattern.
\item for materialized CQ: when no placeholder is defined for a CQ, require that there exists (potentially in a CQ set for different ontology) a different CQ that also produces the same pattern candidate. In that case - there is more than one occurrence of a pattern candidate so it is accepted as a pattern. If pattern candidate is not produced by any other CQ - it is rejected at that stage, so further analysis do not cover these cases.
\end{itemize}





\paragraph{Pattern semantics}

The pattern extraction procedure produced patterns that are semantically the same differing only in minor  aspects like plural vs singular ``be'' verb, using synonyms or using words that could be removed from the CQ without changing its meaning.
For example, consider the patterns
\begin{enumerate}
    \item \textit{Is there EC1 for EC2?} (pattern 26)
    \item \textit{Are there any EC1 for EC2?} (pattern 58)
\end{enumerate}
They are simply different surface realisations of the same question. Then, if we produce CQs by setting the same EC1 and EC2 value respectively for each pattern, we obtain CQs that are semantically the same and thus answers provided by both of them will be the same.

Additionally, one may argue about the detection of chunk boundaries. Consider CQ from DemCare: DemCare\_CQ\_51: \emph{What data are measured for neuromuscular impairment in speech production mechanism?}. After performing the chunk identification procedure as described in the previous paragraph, it would produce the pattern candidate \emph{What EC1 PC1 EC2 in EC3} with: \emph{data} identified as \emph{EC1}, \emph{neuromuscular impairment} as \emph{EC2}, \emph{speech production mechanism} as \emph{EC3} and \emph{are measured for} as \emph{PC1}. Note that two ECs are separated with a preposition, there is no rule that allows a preposition to be inside EC.

Although the ontology can have separate classes for \emph{neuromuscular impairment} and \emph{speech production mechanism}, it also may have one class representing \emph{neuromusular impairment in speech production mechanism}. It is a modelling decision that ontology engineer has to make to select how to model it, so at the level of linguistic structure analysis, we cannot be sure if, for a given pattern candidate,  \emph{What EC1 PC1 EC2 in EC3} or rather \emph{What EC1 PC1 EC2} with EC2 merging EC2 and EC3  would be the actual underlying pattern. Thus, we should also analyze patterns that have ``wide'' ECs, with all ECs separated with pronouns merged into one EC.

Because of those issues, we propose a two step aggregation procedure that joins semantically the same patterns and unify all cases, where it is more than one way of interpretation of chunk boundaries:

\begin{itemize}
\item Normalize words observed in a pattern according to Table \ref{tab:normalizedform}.
\begin{table}[h!]
     \caption{Normalization of words into common forms. REMOVED means that given text is deleted from pattern\label{tab:normalizedform}.}
    \begin{tabular}[t]{ll}
        \hline
        Textual pattern & Normalized form\\
        \hline
        are & is\\
        any & REMOVED\\
        did & do\\
        we & I\\
        does & do\\
        which of & which\\
        has & have\\
        which kind & what kind\\
        will & is\\
        Which (at sentence beginning) & What\\
        possible & REMOVED\\
        are there & REMOVED\\
        \hline
    \end{tabular}
    \end{table}
 That step changes plurals that are observed in patterns into singular forms, remove unimportant words and change less frequent keywords into more popularly used alternatives.

 \item  Replace multiple ECs separated by a preposition using a single EC identifier (Table~\ref{tab:replace}).
    \begin{table}[h!]
     \caption{Replacing complex entity expressions with single identifier\label{tab:replace}.}
        \begin{tabular}[t]{cc}
        \hline
        Textual pattern & Normalized form\\
        \hline
        EC for EC & EC\\
        EC of EC & EC\\
        EC in EC & EC\\
        EC with EC & EC\\
        EC from EC & EC\\
        \hline
    \end{tabular}
\end{table}
\end{itemize}

\subsection{Results}

\begin{table*}[t]
\begin{center}
\caption{Number of pattern candidates and actual patterns}
\label{table:cq_pattern_cand_count}
\centering
 \begin{tabular}{p{.08\textwidth}
 		>{\raggedleft\arraybackslash}p{.08\textwidth}
 		>{\raggedleft\arraybackslash}p{.06\textwidth}
 		>{\raggedleft\arraybackslash}p{.06\textwidth}
 		>{\raggedleft\arraybackslash}p{.09\textwidth}
 		>{\raggedleft\arraybackslash}p{.10\textwidth}
 		>{\raggedleft\arraybackslash}p{.12\textwidth}
 		>{\raggedleft\arraybackslash}p{.12\textwidth}}
 \hline
  Ontology's CQ set & Pattern Candidates & Patterns & Distinct Patterns & CQs covered by patterns & Materialized CQs & Dematerialized CQs & Distinct Higher level patterns\\ 
 \hline\hline
 SWO & 88 & 88 & 72 & 100\% & 1 & 87 & 60 \\
 \hline
 Stuff & 11 & 7 & 6 & 63.6\% & 4 & 7 & 5 \\
 \hline
 AWO & 14 & 10 & 9 & 71.4\% & 6 & 8 & 8\\
 \hline
 Dem@Care & 107 & 90 & 18 & 84.1\% & 107 & 0 & 15\\
 \hline
 OntoDT & 14 & 14 & 8 & 100\% & 0 & 14 & 4\\
  \hline
 Total & 234 & 209 & 106 & 89.3\% & 118 & 116 & 81\\
 \hline
\end{tabular}
\end{center}
\end{table*}

\subsubsection{Linguistic patterns of CQs}


The full list of detected {\em distinct} patterns are included in the appendix as well as on github as part of the dataset. 
The vast majority of CQs share their linguistic structure with them. 
Table \ref{table:cq_pattern_cand_count} provides an overview of pattern distribution among the CQ sets of the ontologies as well as for the whole dataset.
It contains 4 aggregates summarizing each ontology and the whole dataset:
\begin{enumerate}
\item Pattern Candidates -- the number of pattern candidates constructed from CQs for a given ontology. In each case it is equal to number of CQs because we can construct a pattern candidate out of each CQ.
\item Patterns -- the number of pattern candidates that are interpreted as patterns according to the procedure defined in \ref{filtering_sect}.
\item Distinct Patterns -- the number of distinct patterns detected for given ontology. Because multiple patterns can have the same form, distinct patterns tells us how intensively patterns are reused for a given ontology.
\item CQs covered by patterns -- the percentage of CQs for which patterns are detected (having linguistic structure that is recurring among multiple CQs).
\item Materialized CQs -- the number of materialized CQs (without placeholders).
\item Dematerialized CQs -- the number of dematerialized CQs (containing a placeholder).
\end{enumerate}
While there are 116 dematerialised CQs, which may suggest there would be more than 116 patterns, for each dematerialised CQ counts as a pattern, there are several dematerialised CQs that have the same pattern, such as swo82 \textit{What graphics card does [this software] require?} and swo84 \textit{What platform does [the software] run on?} that are of the single pattern \textit{What EC1 PC1 EC2 PC1?}. Due to this, the overall number of unique patterns can be, and is, is less. Notwithstanding, the number is substantially higher than the numbers of patterns reported in related works (19 and 14 in \cite{DBLP:conf/esws/RenPMPDS14,Bezerra14}, respectively), which is at least partially due to the multiple surface realisations for the same question, and therefore we turn to the higher-level patterns in the next subsection.




\subsubsection{Higher level of patterns}

After the generalisation steps, we obtain simpler, higher level, patterns, such as \emph{Is there EC1} that is the simplified version of the following three patterns:
\begin{enumerate}
\item \emph{Are there EC1 in EC2}
\item \emph{Is there EC1 for EC2}
\item \emph{Is there EC1 with EC2}
\end{enumerate}
and likewise for \emph{What is EC1}, which is the base form for the following variants
\begin{enumerate}
\item \emph{What is EC1}
\item \emph{What is EC1 of EC2}
\item \emph{What are EC1}
\item \emph{Which are EC1 of EC2}
\item \emph{What is EC1 of EC2 for EC3}
\item \emph{What are EC1 for EC2}
\item \emph{What is EC1 for EC2}
\end{enumerate}

The process of generating higher level patterns reduces the number of observed patterns, as shown in 
the last column of 
Table~\ref{table:cq_pattern_cand_count}, 
where the normalised patters are indicated with an asterisk in the appendix.  


The process of producing higher level patterns decreased the total amount of distinct patterns by 25. The reduction was mainly observed among patterns detected in datasets of SWO (reduction by 12 patterns), OntoDT (reduction by 4 patterns) and Dem@Care (reduction by 3 patterns). Also, in case of Dem@Care, because it does not utilize placeholders inside CQs, it occurs that new patterns are observed, since very similar single occurrences of pattern candidates are unified into more numerous groups of higher level patterns. Thus, the number of CQs covered by patterns increased for Dem@Care from 84.1\% to 92.5\%.

\subsubsection{Pattern reuse}

 \paragraph{Pattern reuse between CQ sets}
Having detected patterns it is interesting to check if they generalize between CQ sets. Tables \ref{tab:shared_cq_patterns} and \ref{tab:shared_cq_patterns_lv3} lists all patterns occurring in more than one CQ set and notes in which CQ sets the pattern was observed.


\begin{table}[]
    \centering
    \caption{Patterns that are shared by CQ sets of multiple ontologies.}
    \begin{tabular}{ll}
        \hline
        Pattern & In CQ sets for ontologies \\
        \hline
        \hline
        What EC1 PC1 EC2 & SWO, Dem@Care \\
        \hline
        Which EC1 PC1 EC2 & SWO, AWO \\
        \hline
        What are EC1 for EC2 & SWO, OntoDT \\
        \hline
        What is EC1 for EC2 & SWO, OntoDT \\
        \hline
        What is EC1 of EC2 & SWO, AWO \\
        \hline
        Which EC1 are EC2 & Dem@Care, AWO \\
        \hline
    \end{tabular}
    \label{tab:shared_cq_patterns}
\end{table}

\begin{table}[]
    \centering
    \caption{Higher Level Patterns that occur in more than one CQ set.}
    \begin{tabular}{ll}
        \hline
        Pattern & In CQ sets for ontologies \\
        \hline
        \hline
        What type of EC1 is EC2 & SWO, Stuff, Dem@Care \\
        \hline
        What EC1 PC1 EC2 & SWO, Dem@Care, AWO \\
        \hline
        What is EC1 & SWO, OntoDT, Dem@Care \\
        \hline
        What EC1 PC1 I PC1 EC2 & SWO, AWO \\
        \hline
        Is EC1 EC2 & SWO, AWO \\
        \hline
        Is there EC1 & SWO, AWO \\
        \hline
        What EC1 PC1 EC2 PC1 & SWO, AWO \\
        \hline
        What EC1 is EC2 & Dem@Care, AWO \\
        \hline
    \end{tabular}
    \label{tab:shared_cq_patterns_lv3}
\end{table}

It is interesting that there is no pattern shared among all CQ sets. Even considering the higher level pattern detection, the maximum number of ontologies sharing the same pattern is 3 out of 5. That means that the linguistic forms of CQs defined against different ontologies are highly different. Out of 106 patterns, only 6 are shared among more than one CQ set, and out of 81 higher level patterns, still only 8 of them are shared. The reason for that may be the fact that there are no good practices proposed on how to construct CQs, so different domain experts and ontology engineers use different forms to state their CQs, or perhaps there is a domain dependence in the types of queries.

\paragraph{Pattern reuse inside ontologies}

\begin{figure}[h]
  \centering
    \includegraphics[width=1.0\linewidth]{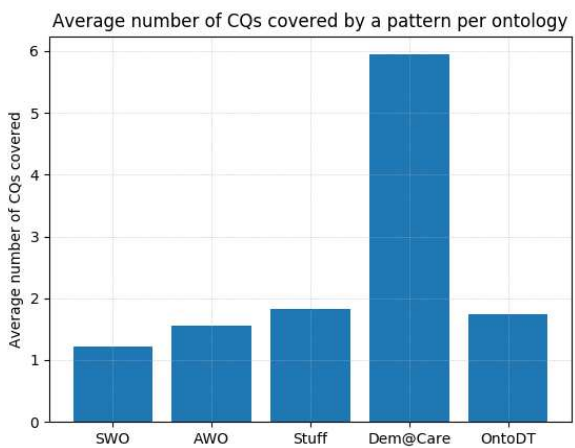}
  \caption{Average CQs covered by a single pattern per given CQ set of an ontology.}
  \label{fig:barplot}
\end{figure}

\begin{figure}[h]
  \centering
    \includegraphics[width=1.0\linewidth]{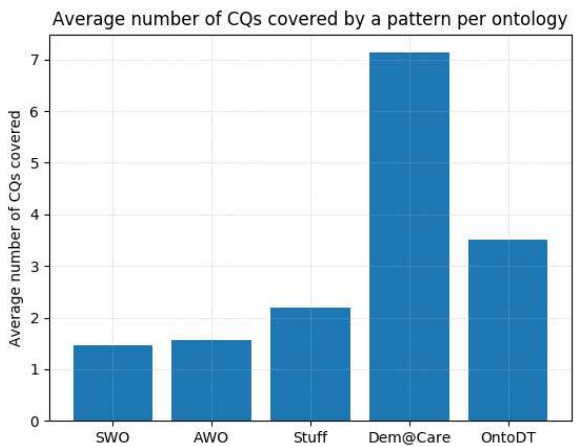}
  \caption{Average CQs covered by a single higher level pattern per given CQ set of an ontology.}
  \label{fig:barplot_lv3}
\end{figure}

Figures \ref{fig:barplot} and \ref{fig:barplot_lv3} present an average number of CQs that are covered by patterns and higher level patterns respectively for a given ontology dataset. In both cases, on average all patterns cover more than one CQ. The highest value can be observed for Dem@Care. The reason for that is the fact that CQs defined for Dem@Care do not utilize placeholders. Instead multiple similar CQs are produced that produce same pattern.

But also for datasets where all CQs utilize placeholders, some reusage can be observed. For OntoDT, on average 2 CQs are covered by a single pattern. When considering higher level patterns, each higher level pattern on average covers 3.5 (dematerialised) CQs.

We can suppose that if ontologies had defined all CQs without utilizing placeholders, the reusage for each of them would be as high as for Dem@Care. The idea of placeholder was introduced to propose a common linguistic form of a CQ that can be used multiple times. Thus, we can assume that in general ontology engineers tend to reuse the patterns.

\subsection{Concluding remarks}

The presence of placeholders in 4 out of 5 datasets, high reusage of patterns in the dataset without placeholders and the fact that only few patterns are shared among different datasets lead to a conclusion that,
in practice, there will be a high reuse of the same CQ patterns, albeit not at the level of CQ specification.
The great expressivity of natural language leads different experts to formulate CQs in different ways, thus the general forms of CQs are not shared among different datasets.



\section{Generating SPARQL-OWL queries from CQs}
\label{sec:cq:sparql}
In this section, we carry out and examine the `translation' of CQs to a form that can be evaluated against an ontology. 

As first preliminary observation, we observe that an OWL ontology can be serialized as an RDF/XML graph \cite{owl-rdf-mapping} and thus queried using SPARQL Query Language  \cite{sparql}. 
In its base form SPARQL is basically a pattern matching language and as such does not provide any reasoning capabilities; however, it is possible to introduce these by using SPARQL Entailment Regimes \cite{sparql-entailment-regimes}.
In particular, we employ OWL 2 Direct Semantics Entailment Regime.
Intuitively, it allows us to construct a SPARQL query such that its WHERE clause contains OWL axioms, possibly with some of its IRIs and literals replaced by SPARQL variables.
The results of the execution of such a query are all the variable mappings such that the axioms obtained by applying these mapping to the axioms in the query, are entailed by the queried ontology.
SPARQL, being a query language for RDF, employs Turtle syntax \cite{turtle} to express Basic Graph Patterns (BGPs) and this convention is kept also for expressing OWL axioms, i.e., their RDF representation is used \cite{owl-rdf-mapping}.
This is consistent with how the only available implementation behaves\footnote{\url{https://github.com/iliannakollia/owl-bgp}} \cite{DBLP:conf/esws/KolliaGH11,DBLP:journals/jair/KolliaG13}.

The second preliminary comment is that we note that, unlike Dennis et al.~\cite{DBLP:conf/semweb/DennisDDP17}'s claim, CQs {\em do not} have to have specific presuppositions other than vocabulary, but queries do, for it is the queries that are specific to the ontology and the modelling style used and other modelling decisions made. We can make this distinction here, because of the separation of concerns between the linguistics of the CQs on the one hand and the queries and ontology how it it realised on the other hand, rather than having the two combined as in \cite{Bezerra14,DBLP:conf/esws/RenPMPDS14,DBLP:conf/semweb/DennisDDP17}.

To obtain the formal representation of gathered CQs we manually translated them to SPARQL-OWL.
The process of translating a CQ was organized as follows.
First, we identified keywords in the CQ, which then were used to identify relevant vocabulary in the ontology.
If we could not find a match for a given keyword, we looked for different surface forms and words with similar meaning.
Next we identified a subset of expected answers for the query, either by identifying relevant vocabulary using the answers provided with the CQ, or by trying to answer the CQ using the knowledge present in the ontology.
Then, we used the identified vocabularies to decide how to construct the query in such a way, that it would provide the expected answers.
This step required careful inspection of the modelling style of the ontology.
Finally, the query was constructed and tested using OWL-BGP, to verify whether it was constructed correctly and yields the expected answers.

During the process we observed a high variablity in the structure of the resulting queries and we identified two main causes: (i) the CQs themselves can vary significantly within and between ontologies; (ii) knowledge in the ontology can be modelled in various ways, sometimes highly diverging from the "default" approaches suggested by the OWL standards.
Below, we discuss the results in details, providing justification for the decisions made and highlighting the most interesting cases.

In the remainder of the paper, we present and discuss multiple SPARQL-OWL queries.
To keep them easily readable, we use prefixes to replace full namespaces in URIs, but for brevity, we omit the preambles of the queries.
Instead, we provide the full list of the prefixes used and their corresponding namespaces in Table \ref{tab:prefixes}.

\begin{table*}
	\caption{Prefixes used thruought the paper in the SPARQL-OWL queries \label{tab:prefixes}
	}
	\centering
	\begin{tabular}{ll}
		\hline
		prefix & namespace \\
		\hline
		\hline
		\texttt{rdf:} & \url{http://www.w3.org/1999/02/22-rdf-syntax-ns#} \\
		\texttt{rdfs:} & \url{http://www.w3.org/2000/01/rdf-schema#} \\
		\texttt{owl:} & \url{http://www.w3.org/2002/07/owl#} \\
		\texttt{xsd:} & \url{http://www.w3.org/2001/XMLSchema#} \\
		\hline
		\texttt{awo:} & \url{http://www.meteck.org/teaching/ontologies/AfricanWildlifeOntology1.owl#} \\
		\hline
		\texttt{stuff:} & \url{http://www.meteck.org/files/ontologies/stuff.owl#} \\
		\hline
		\texttt{event:} &  \url{http://www.demcare.eu/ontologies/event.owl#} \\
		\texttt{exch:} & \url{http://www.demcare.eu/ontologies/exchangemodel.owl#} \\
		\url{home:} & \url{http://www.demcare.eu/ontologies/home.owl#} \\
		\texttt{lab:} & \url{http://www.demcare.eu/ontologies/lab.owl#} \\
		\hline
		\texttt{swo:} & \url{http://www.ebi.ac.uk/swo/} \\ 
		\texttt{efo-swo:} & \url{http://www.ebi.ac.uk/efo/swo/} \\
		\texttt{maturity:} & \url{http://www.ebi.ac.uk/swo/maturity/} \\
		\texttt{interface:} &
		\url{http://www.ebi.ac.uk/swo/interface/} \\		
		\texttt{obo:} & \url{http://purl.obolibrary.org/obo/} \\
		\hline
		\texttt{OntoDT:} & \url{http://www.ontodm.com/OntoDT#} \\
		\texttt{OntoDT2} & \url{http://ontodm.com/OntoDT#} \\
		\hline
	\end{tabular}
\end{table*}

\subsection{African Wildlife Ontology}

For \emph{African Wildlife Ontology} (AWO) we gathered 14 competency questions.
Six of them (CQ5, CQ9--CQ13), concerning drinking, habitats and conservation status, were deemed impossible to translate due to the lack of vocabulary in the ontology.
Four questions (CQ1, CQ6--CQ8) represented the same pattern of asking about classes connected with a property specified in the questions using existential restriction.
For example awo\_6 \emph{Which plants eat animals?} was translated as 
\begin{alltt}
SELECT DISTINCT ?eats
WHERE \{
    ?eats rdfs:subClassOf awo:plant, [
        a owl:Restriction ;
        owl:onProperty awo:eats;
        owl:someValuesFrom awo:animal
    ] .
    FILTER(?eats != owl:Nothing)
\}
\end{alltt}
The query asks for all subclasses of the class \texttt{:plant}, which has existential restriction on the property \texttt{:eats} targeting the class \texttt{:animal}.
Because SPARQL-OWL uses reasoning during answering the query, the answer for the query will contain classes with more specific restrictions (e.g., \texttt{:impala}, which is a subclass of \texttt{:animal}).
The filter clauses was added to remove, technicaly correct, but useless, answer of \texttt{owl:Nothing}, i.e., the bottom concept.

Awo\_2 \emph{Is [this animal] a herbivore?} is a simple question to check whether a class is a subclass of another class, while awo\_4 \emph{Does a lion eat plants or plant parts?} is similar, but yields a more complicated query: the superclass in the query is an existential restriction on the property \texttt{:eats} and \emph{plants or plant parts} must be modeled as an union of classes \texttt{:plant} and \texttt{:PlantParts}.
Awo\_14 \emph{Are there animals that are carnivore but still eat some plants or parts of plants?} can be interpreted in two ways.
One possibility is to treat the question as a question about presence of particular knowledge in the ontology or, in other words, \emph{Does the ontology contain any carnivore that eats some plants or parts of plants?}.
In this interpretation the question resembles the formerly discussed questions awo\_1 and awo\_6--awo\_8.
The other possiblity is to interpret the question as a question about possiblity of existence of such an animal or, in other words, \emph{Is it possible for a carnivore to eat  some plants or parts of plants?}
Then the questions corresponds to the query about (lack of) a disjointness axiom between the class \texttt{:carnivore} and the existential restriction already discussed in awo\_4.
Because the presence of the disjointness axiom corresponds to the negative answer to the original question, the query must contain negation, which can be obtained using \texttt{FILTER NOT EXISTS}.

\subsection{Stuff}

For the \emph{Stuff} ontology, we collected 11 competency questions, that can be divided into 6 distinct categories w.r.t. the required reasoning.
 Stuff\_01 \emph{Is [this stuff] a pure or a mixed stuff?} is a question to decide whether a class (e.g., \texttt{Mayonnaise}) is a subclass of one of the two specified classess \texttt{PureStuff}, \texttt{MixedStuff}.
Stuff\_02 \emph{What is the difference between [this colloid] and [this colloid]?} and stuff\_08 are questions about finding differences between definitions of the classes.
In principle, the ontology contains enough information to answer the questions, as the referred classes have their complete definitions there.
Unfortunately, this is a non-standard reasoning task and is not possible to express it using SPARQL-OWL.
Stuff\_03 \emph{In which phases are the stuffs in [this colloid]?} requires finding all the superclasses of a specified class (e.g., \texttt{emulsion}), that are complex class expression with nested existential restrictions on properties, respectively, \texttt{hasPartStuff} and \texttt{hasState}.
The class expressions from the nested restriction are the answers to the query.
Stuff\_04 \emph{Can a solution be a pure stuff?} and stuff\_06 asks whether two classes are disjoint.
Stuff\_04 requires negation, similarly to stuff\_04 from AWO.
Stuff\_05 \emph{Which kind of stuff are [these stuffs]?}, stuff\_09 and stuff\_10 are questions about all the superclasses of a given class (Sutff\_05: \emph{these stuffs}, e.g., \texttt{emulsion}), that are subclasses of another class (Sutff\_05: \texttt{Stuff}).
Stuff\_11 \emph{Where do I categorise bulk like [this bulk]?} is a simpler version, that does not contain the second requirement
Stuff\_07 \emph{Which stuffs have as part exactly two substuffs?} is a question about all the subclasses of a class expression using cardinality restriction $=2$ on the property \texttt{hasSubStuff}.

\subsection{Dem@Care}

The Dem@Care ontology is accompanied by 107 competency questions and (a subset of) expected answer for each of them.
47 of them lacks the appropriate vocabulary and/or knowledge in the ontology and thus can not be modeled
The remainder can be divided into six groups depending on the shape of the query.

Group I contains three questions (DemCare\_CQs: 4, 6, 8) askings for instances of a given class, e.g., DemCare\_CQ\_4 \emph{What is the gender information?} corresponds to the BGP \texttt{?x rdf:type lab:GenderType}.

Group II contains 17 questions (DemCare\_CQs: 7, 23, 29--32, 57, 65, 72, 75, 77, 78, 82, 87, 105--107) that ask for all the proper subclasses of a given class, e.g., the BGP of DemCare\_CQ\_7 \emph{What types of clinical data are collected?} is
\begin{alltt}
?x rdfs:subClassOf lab:ClinicalAssessment .
FILTER(?x != lab:ClinicalAssessment)
\end{alltt}
 where \texttt{?x} is the distinguished variable and the filter clause is to ensure only proper subclasses are considered.

Group III, containing 7 questions (DemCare\_CQs: 83-85, 88, 90, 98, 104), corresponds to questions asking about the direct subclassess of a given class.
For example DemCare\_CQ\_83 \emph{What are the main types of entities?} corresponds to the following graph pattern:
\begin{alltt}
?x rdfs:subClassOf event:Entity .
FILTER NOT EXISTS \{
    ?x rdfs:subClassOf ?y .
    ?y rdfs:subClassOf event:Entity.
    FILTER(?y != event:Entity && ?x != ?y)
\}
FILTER(?x != event:Entity && ?x != owl:Nothing)
\end{alltt}
We distinguished group II from group III by, respectively, absence or presence the word \emph{main} (c.f., \emph{the main types} above) in the question.

Group IV is by far the largest, containing 24 questions (DemCare\_CQs: 3, 9, 15, 19, 21, 40, 47, 50, 52--54, 58--60, 62--64, 68, 76, 80, 89, 99, 101, 103).
These are questions about relations an object of a given class is expected to have, i.e., about property names present in existential restrictions.
For example, DemCare\_CQ\_3 \emph{What types of demographic data are collected?} can be represented by the following graph pattern:
\begin{alltt}
lab:DemographicCharacteristicsRecord 
  rdfs:subClassOf [
    a owl:Restriction ;
    owl:onProperty ?x;
    owl:someValuesFrom []
].
\end{alltt}
where \texttt{lab:DemographicCharacteristicsRecord} is the class named in the query, and \texttt{?x} is the distinguished variable.
There is also another possiblity to represent questions from this category.
Consider DemCare\_CQs\_89 \emph{What are the main types of information describing an event?} and its gold answers \emph{The agent of the event (i.e. the referred person, object or room), start time, duration, and location (where applicable)}.
The \emph{where applicable} part hints that there are properties that are expected only for some subtypes of an event (i.e., subclasses of the class \texttt{event:Event}).
We can thus consider the following graph pattern
\begin{alltt}
[] rdfs:subClassOf event:Event, [
    owl:onProperty ?p;
    owl:someValuesFrom []
].
\end{alltt}
Here, we consider all the subclasses of the class \texttt{event:Event} and ask for properties they are expected to have.
Queries of such form are more general than the queries of the earlier form in the sense that the latter always returns at least the same information as the earlier (query subsumption).

Group V contains 7 questions (DemCare\_CQs: 33--39) that ask about class names of values for a specified property.
For example, DemCare\_CQs\_33 \emph{What is assessed in the walking task?}	can be represented as a query with the following graph pattern:
\begin{alltt}
lab:S1\_P11\_WalkingTask rdfs:subClassOf [
    a owl:Restriction;
    owl:onProperty lab:measuredData;
    owl:someValuesFrom ?x
].
?x rdfs:subClassOf lab:MeasuredData.
FILTER(?x != lab:MeasuredData)
\end{alltt}
The first triple pattern selects all the class names that are present in a existential restriction for the class \texttt{lab:S1\_P11\_WalkingTask} (or its ancestors) on the property \texttt{lab:measuredData}, while the second along with the filter clause ensure that only proper subclasses of the class \texttt{lab:MeasuredData} are returned.
Should the second part be omitted, the query would return also, e.g, \texttt{owl:Thing}, which certainly does not answer the question.

Group VI contains 2 questions that are classes on their own and require detailed discussion.
DemCare\_CQ\_67 \emph{What information is of clinical interest regarding food and drink preparation?}
is a union of two questions from the group four (questions about property names) with two different classes: \texttt{event:PrepareMeal} and \texttt{event:PrepareDrink}.
It can be realized using SPARQL \texttt{UNION} clause or using OWL \texttt{owl:unionOf}, and the graph pattern of the first possibility is presented below.
\begin{alltt}
\{
    [] rdfs:subClassOf event:PrepareMeal, [
        owl:onProperty ?p;
        owl:someValuesFrom []
    ]. 
\} UNION \{
    [] rdfs:subClassOf event:PrepareDrink, [
        owl:onProperty ?p;
        owl:someValuesFrom []
    ]. 
\}
\end{alltt}

The second question in the group DemCare\_CQ\_100	\emph{What are the main types of data a report may refer to?}.
Its expected answers are \emph{Questionnaires, clinical characteristics, demographic data, \ldots} and they can indeed be found in the axioms describing the class \texttt{exch:Report}, namely in a universal restriction on the property \texttt{exch:describes}.
To retrieve them all it is necessary to dig into the RDF list representation of the union, as presented in the BGP below.
\begin{alltt}
exch:Report rdfs:subClassOf [
    a owl:Restriction ;
    owl:onProperty exch:describes ;
    owl:allValuesFrom [
        a owl:Class ;
        owl:unionOf/rdf:rest*/rdf:first ?c
] ].
\end{alltt}

\subsection{Software Ontology}

The Software Ontology SWO is accompanied by the set of 88 competency questions.
The quality of the provided questions varies greatly, from very vague (e.g., swo12 \emph{What software works best with my dataset?}) to very specific ones (e.g., swo22 \emph{Can software A work with data that are output from software B?}).
It must be noted that almost none of the questions are context-independent, in this sense that they tend to contain unresolvable pronouns (e.g., \emph{this software} in swo20 \emph{What is the valid input for this software?}), placeholders (e.g., A and B in swo22, above).
The sole exception is swo18 \emph{What software can read a .cel file?}.
We performed the translation introducing placeholders in the queries to represent the placeholders/pronouns from the question.

Out of these 88 questions, 46 were deemed impossible to translate to SPARQL-OWL, because lack of the vocabulary in the ontology or due to their ambiguity: swo5--6, swo12--13, swo17, swo23--24, swo27--28, swo30, swo32--34, swo37--38, swo40--43, swo46--53,  swo55--56, swo61, swo63--64, swo66, swo68-69, swo71, swo74--75, swo77, swo79--82, swo84--87.

The remaining questions can be divided into 9 groups based on the shape of the graph pattern in the final query.
The division is coarse and aimed at underlying recurring patterns rather than at introducing a strict classification.
As mentioned earlier, the questions contain placeholders and we denote them in the graph patterns using variables prefixed with \texttt{\$} instead of \texttt{?}.

Group I consists of questions about classes occuring in an existential restriction for a given class.
For example, consider the question swo04 \emph{Which of the named and published "algorithms" does this tool use?}.
This can be achieved with the following graph pattern:
\begin{alltt}
\$sw rdfs:subClassOf [
    owl:onProperty efo-swo:SWO_0000740;
    owl:someValuesFrom ?alg
] .
?alg rdfs:subClassOf obo:IAO_0000064 .
FILTER(?alg != obo:IAO_0000064 && 
       ?alg != owl:Nothing) .
\end{alltt}
\texttt{SWO\_0000740} is an object property labeled \emph{implements} and its range is the class \texttt{obo:IAO\_0000064} \emph{algorithm}.
The second subclass expression ensures that only algorithms are listed, otherwise also superclasses of the class \texttt{obo:IAO\_0000064} would be returned.
The filter expression is added to remove this particular class and the bottom concept \texttt{owl:Nothing} from the results, as they are both meaningless.
CQs with similar form of the graph pattern are: swo07, swo45, swo83.

Group II is very similar, but requires retrieving an actual value (an individual or a literal) rather than a class name.
Consider swo36 \emph{What is the homepage of the software?} which can be represented using the following graph pattern:
\begin{alltt}
\$sw rdfs:subClassOf [    
    owl:onProperty swo:SWO_0004006 ;
    owl:hasValue  ?url
] .
\end{alltt}
The property \texttt{SWO\_0004006} is a data property \emph{has website homepage}.
The main difference between this graph pattern and the previous one is the usage of \texttt{owl:hasValue} instead of \texttt{owl:someValuesFrom} and the lack of the filter expression, which is not needed in this situation.
The other questions following the same pattern are: swo29 (approximation), swo31, swo39, swo54 (uses an object property instead of a data property), swo70, swo72, swo73.
An interesting corner case in this grup is swo76 \emph{Is there a publication with it?}.
The ontology itself does not provide appropriate vocabulary and in general this is impossible to answer.
However, the property \texttt{SWO\_0000043} \emph{has documentation} is hacked in an interesting way: most of its usages in the ontology has values being URLs starting with \url{http://dx.doi.org/}, thus refering to a publication with a DOI.
An approximate answer to this particular question could be thus the following graph pattern:
\begin{alltt}
\$sw rdfs:subClassOf [
    owl:onProperty swo:SWO_0000043 ;
    owl:hasValue ?doc
].
FILTER(STRSTARTS(?doc, "http://dx.doi.org/"))
\end{alltt}
To answer literally for the posed query the pattern should be used in an ASK query, but it can be as well used in a SPARQL query to produce a list of publications instead.
Another interesting variant is the question swo44 \emph{How long has this software been around?} which requires postprocessing the answer.
Consider the following graph pattern:
\begin{alltt}
\$sw rdfs:subClassOf [
    owl:onProperty maturity:SWO_9000068 ;
    owl:hasValue ?date
]

BIND(now()-xsd:dateTime(?date) AS ?result)
\end{alltt}
This is a pattern typical for the questions in this group plus a bind expression to actually compute the difference between the release date and the current time, which is the expected result of the query.

Group III is constituted by questions of type \emph{Which software \ldots?}, e.g., swo08 \emph{Which software can perform task x?}.
This is also similar to Group I, but the placeholder is placed in the superclass expression and the distinguished variable is the subclass position.
The sample question can be realized with the following graph pattern:
\begin{alltt}
?sw rdfs:subClassOf swo:SWO_0000001 , [
    owl:onProperty swo:SWO_0040005 ;
    owl:someValuesFrom \$task
] .
\end{alltt}
The class \texttt{SWO\_0000001} \emph{software} is in the query to ensure that only actual pieces of software are returned, while the existential restriction on the property \texttt{SWO\_0040005} \emph{is executed in} ensures that the additional condition of the question (i.e., performing task x) is fulfiled.
The query can again be extended with filtering out the bottom concept.
The other question in this group is swo14.

Group IV consists of yes-no questions of type \emph{Is this software\ldots}, e.g., swo53 \emph{Is this software available as a web service?}.
This is a variant of Group III, because the only difference is in using a placeholder instead of the normal variable in the graph pattern and the appropriate SPARQL verb is ASK rather than SELECT.
The sample question can be realized using the following graph pattern:
\begin{alltt}
\$sw rdfs:subClassOf [
    owl:onProperty swo:SWO_0004001 ;
    owl:someValuesFrom interface:SWO_9000051
] .
\end{alltt}
The property \texttt{SWO\_0004001} is labeled \emph{has interface} and the class \texttt{SWO\_9000051} \emph{web service}.
The other questions in the group are: swo9, swo65
Also swo88 \emph{Do I need a license key to use it?} can be approximated by this grup.
In general the question is about a technical side of licensing and can not be represented using the available vocabulary, but one can approximate it as a question about whether the licence is propertiary or not.

Group V consists of conjuncive mixtures of the previous groups.
For example, consider swo11 \emph{Which visualisation software is there for this data and what will it cost?}.
The ontology does not cover the area of software costs, but the remainder of the question, i.e., \emph{Which visualisation software is there for this data?} can be answered using an conjunction of two pattern from Group III.
In other words, the question can be seen as a conjunction of two questions: \emph{Which visualisation software is there?} and \emph{Which software is there for this data?}.
This yields the following graph pattern:
\begin{alltt}
?sw rdfs:subClassOf swo:SWO_0000001 ;
    rdfs:subClassOf [
        # has specified data input        
        owl:onProperty swo:SWO_0000086 ;
        owl:someValuesFrom ?data
    ] ;
    rdfs:subClassOf [
        # is executed in
        owl:onProperty swo:SWO_0040005 ;    
        # data visualisation
        owl:someValuesFrom efo-swo:SWO_0000724
    ] ;
FILTER(?sw != owl:Nothing)
\end{alltt}
The other similar questions are swo15, swo57 (two patterns from Group II, one with a variable replaced by a placeholder)

Group VI are the questions that require a nested existential restriction, possibly in a conjunction with patterns from the previous groups.
For example, consider swo01 \emph{What is the algorithm used to process this data?}
The question is vague in this sense that, in principle, it is possible to use a sorting algorithm on an arbitrary binary data, but such an operation is usually meaningless and thus such an answer would be unexpected.
Moreover, the ontology deals with the algorithms only to this extent that algorithms are implemented by something, but they are not described by themselves, in particular: their expected inputs and outputs are unknown.
Finally, we assumed that \emph{this data} referes to a data format.
Under these assumptions the question could be rewritten as follows: \emph{What are the algorithms implemented by software that is capable of processing data in this format?}, \emph{this format} being a placeholder.
This yielded quite a complex query presented below.
In lines 1--15 the query looks for a piece of software (line 1, the class \texttt{swo:SWO\_0000001} \emph{software}) capable of processing an input (lines 2--10, an existential restriction on the property \texttt{swo:SWO\_0000086} \emph{has specified data input}) such that it is data (line 5, the class \texttt{obo:IAO\_0000027} \emph{data}) and it is expressed in the given format (lines 6--7, the existential restriction on the property \texttt{swo:SWO\_0004002} \emph{has format specification}).
From this software, the implemented algorithms are extracted in lines 11-15 (the existential restriction on the property \texttt{SWO\_0000740} \emph{implements}.
As we are interested only in the actual algorithms further filtering is performed in lines 16--17 by ensuring that the variable \texttt{?alg} is bound to a proper subclass of the class \texttt{obo:IAO\_0000064} \emph{algorithm}.

\begin{lstlisting}
[] rdfs:subClassOf swo:SWO_0000001 ; 
  rdfs:subClassOf [       
    owl:onProperty swo:SWO_0000086 ;
    owl:someValuesFrom [
      owl:intersectionOf (obo:IAO_0000027 [
        owl:onProperty swo:SWO_0004002 ;
        owl:someValuesFrom $format
      ] )
    ]
  ] ;
  rdfs:subClassOf [
    owl:onProperty efo-swo:SWO_0000740;
    owl:someValuesFrom ?alg
] .
?alg rdfs:subClassOf obo:IAO_0000064 .
FILTER(?alg != obo:IAO_0000064) .
\end{lstlisting}
Similar questions are swo18--21, swo58, swo62, swo78.

Group VII are the questions that require comparing (in a broad sense) two entities on a specified criterion.
For example consider swo02 \emph{What are the alternatives to this software?}.
Again, this question can not be directly answered by the ontology, so we assumed that a software can be treated as an alternative for another software if there exists an algorithm implemented by both pieces of software.
We obtained the following BGP:
\begin{lstlisting}
$sw rdfs:subClassOf [
  owl:onProperty efo-swo:SWO_0000740;
  owl:someValuesFrom ?alg
] .
?alg rdfs:subClassOf obo:IAO_0000064 .
FILTER(?alg != obo:IAO_0000064) .
?alt rdfs:subClassOf swo:SWO_0000001, [
  owl:onProperty efo-swo:SWO_0000740;
  owl:someValuesFrom ?alg
] .
FILTER($sw != ?alt)
\end{lstlisting}

For the given software \texttt{\$sw} we extract the algorithms using the same approach as in the query for swo01 (lines 1--6).
We then look for pieces of software that implement any of these algorithms (lines 6--10) and ensure and that we provide an actual alternative, i.e., a different piece of software in line 11.
Other questions in the group are swo03 (a semantic duplicate of swo02), swo16, swo22.

Group VIII are the questions that required disjunction (i.e., \texttt{owl:unionOf} or SPARQL \texttt{union}) in their corresponding query.
There were three such questions and we discuss reasons and usage of disjunction in them, but for sake of brevity, we abstain from providing them in the text.
In the first of them, swo25 \emph{What open source, maintained software can I use to process these in this format?} the word \emph{maintained} can actually be mapped to two terms from SWO: \texttt{SWO\_9000065} \emph{Live} and \texttt{SWO\_9000073} \emph{Maintained}, so we used \texttt{owl:unionOf} to include both.
In swo59 \emph{What license does it have and what is its permissiveness?} we used \texttt{union} join two cases: the case with a license without any clauses describing it (i.e., without information about its permissiveness available), and with them.
Finally, in swo67 \emph{Is it free or not?} we use \texttt{union} to combine three separate license clauses representing free software: \texttt{swo:SWO\_9000030} \emph{usage unrestricted}, \texttt{swo:SWO\_9000020} \emph{source code available} and \texttt{SWO\_1000059} \emph{free}.

Group IX contains a single question swo26 \emph{Is the output format of it proprietary?}.
The ontology does not deal with licenses of the data formats, so in principle the question can not be answered.
We provide quite a complex proxy based on an assumption that a data format is proprietary if there is no open-source software capable of producing it.
We approached it by using SPARQL \texttt{FILTER NOT EXISTS} clause with a BGP from Group VI inside.
The interested reader is referred to the dataset for the full code of the query.

\subsection{OntoDT}

OntoDT is equipped with the set of 14 competency questions.
One of them, ontodt\_09, is a instance-level question and thus we did not translate it.
In the remainder, most of them are concerned with classes related to each other by some property.
For example, the question ontodt\_01 \emph{What is the set of characterizing operations for [a datatype X]?} was translated as 
\begin{alltt} 
$X$ rdfs:subClassOf  OntoDT2:OntoDT_487147, [
  a owl:Restriction ; 
  owl:onProperty OntoDT:OntoDT_0000400 ; 
  owl:someValuesFrom ?x
] 
\end{alltt}
where \texttt{\$X\$} corresponds to the placeholder \emph{[a datatype X]} and \texttt{?x} is the result.
The same pattern is exhibited by ontodt\_02.
Similar questions are: ontodt\_03 and ontodt\_10--ontodt\_14, where we check whether \texttt{?x} is a subclass of a specified class, and ontodt\_04--ontodt\_05, where \texttt{\$X\$} and \texttt{?x} swapped places, i.e., we ask about subclasses of a particular class expression.

The question ontodt\_06	\emph{What is the set of datatypes that have [a datatype quality X] and [characterizing operation Y]?} also exhibits a similar pattern, but twice: we demand that the resulting classes are subclasses of two class expressions with an existential restriction.

For question ontodt\_07 \emph{What are the aggregated datatypes that have [an aggregate generator property X]?} the resulting query is more complex, because we need to query for subclasses of an existential restriction with a nested existential restriction:
\begin{alltt}
?x  rdfs:subClassOf OntoDT2:OntoDT_378476, [ 
  a owl:Restriction ; 
  owl:onProperty OntoDT:OntoDT_0000405;    
  owl:someValuesFrom [ 
    a owl:Restriction ; 
      owl:onProperty obo:OBI_0000298;    
      owl:someValuesFrom $X$
  ]
] . 
$X$ rdfs:subClassOf OntoDT2:OntoDT_283020 . 
\end{alltt}
In this BGP \texttt{\$X\$} corresponds to the placeholder in the question \texttt{[an aggregate generator property X]}, while \texttt{?x} is the distinguished variable.
The query corresponding to the question ontodt\_08 is very similar.

\section{SPARQL-OWL query set analysis}
\subsection{General overview of the dataset}
In total, 131 out of 234 competency questions can be translated into a SPARQL-OWL form. The table \ref{table:cq_sparql_translatability} summarizes how many CQs can be translated for each ontology. The reason why some CQs are not translated is due to missing vocabulary in the ontology to construct expected query or expressing the CQ in a too vague way.

\begin{table}[h]
\begin{center}
\caption{Translatability of competency questions}
\label{table:cq_sparql_translatability}.

\centering
 \begin{tabular}{l>{\raggedleft\arraybackslash}p{.2\linewidth}>{\raggedleft\arraybackslash}p{.2\linewidth}} 
 \hline
 Ontology Name & CQ count & Translated CQ count \\
 \hline\hline
 Software ontology & 88 & 42\\ 
 \hline
 Stuff Ontology& 11 & 9\\
 \hline
 African Wildlife Ontology & 14 & 7\\
 \hline
 Dem@Care & 107 & 60\\
 \hline
  OntoDT & 14 & 13\\
  \hline
\end{tabular}
\end{center}
\end{table}

\subsection{SPARQL-OWL keyword frequency analysis}
Table \ref{table:sparql_keywords} contains a list of keywords with number of queries they were used in and ontology names in which those keywords were identified. The maximum value that the second column can have is equal to the number of questions with SPARQL-OWL query defined (131). Among collected ontologies existential restriction and subclassing is used in most of queries, while there is very little usage of cardinality restrictions and universal restrictions. Separating SPARQL-OWL queries to groups of queries stated against the same ontology, keywords: where, select, rdf:type\/a, rdfs:subClassOf, someValuesFrom and onProperty are occuring in some queries in every group.
\begin{table*}[h]
\begin{center}
\caption{Keywords usage among SPARQL-OWL queries}
\label{table:sparql_keywords}.

 \begin{tabular}{lrl} 
 \hline
 Keyword & Count & Occurences in ontologies \\
 \hline\hline
 WHERE & 131 & Dem@Care (60), SWO(42), OntoDT(13), Stuff(9), AWO(7) \\
 \hline
 rdfs:subClassOf & 125 & Dem@Care(57), SWO(42), OntoDT(13), Stuff(7), AWO(6) \\
 \hline
 SELECT & 114 & Dem@Care(60), SWO(30), OntoDT(13), Stuff(7), AWO(4)\\
 \hline
 owl:onProperty & 96 & SWO(42), Dem@Care(33), OntoDT(13), Stuff(2), AWO(6)\\
 \hline
 owl:someValuesFrom & 83 & SWO(31), Dem@Care(32), OntoDT(13), AWO(6), Stuff(1)\\
 \hline
 rdf:type \/ a & 72 & SWO(40), OntoDT(13), Dem@Care(11), AWO(6), Stuff(2) \\
 \hline
 DISTINCT & 71 & Dem@Care(57), Stuff(6), SWO(4),  AWO(4)\\
 \hline
 owl:restriction & 69 & SWO(40), OntoDT(13), Dem@Care(8), AWO(6), Stuff(2)\\
 \hline
 FILTER & 58 & Dem@Care(31), SWO(16), Stuff(6), AWO(5) \\
 \hline
owl:Nothing & 34 & SWO(6), AWO(4), Dem@Care(24)\\
 \hline 
 ASK & 17 & SWO(12), Stuff(2), AWO(3)\\
 \hline
 owl:hasValue & 13 & SWO(13)\\
 \hline
 NOT EXISTS & 11 & Dem@Care(7), SWO(2), Stuff(1), AWO(1)\\
 \hline
 owl:intersectionOf & 7 & SWO(7)\\
 \hline
 owl:unionOf & 4 & AWO (2), Dem@Care(1), SWO(1)\\
 \hline
 UNION & 3 & SWO(2), Dem@Care(1)\\
 \hline
 owl:disjointWith & 3 & Stuff(2), AWO(1)\\
 \hline 
 owl:allValuesFrom & 1 & Dem@Care(1) \\
 \hline
 owl:cardinality & 1 & Stuff(1) \\
 \hline
 rdf:first & 1 & Dem@Care(1) \\
 \hline
 rdf:rest & 1 & Dem@Care(1) \\
 \hline
\end{tabular}
\end{center}
\end{table*}

\subsection{Recurring patterns in the WHERE clauses}

SPARQL-OWL allows for a lot of flexibility in writing queries.
The reasons are, among others:
\begin{itemize}
\item Abbreviated notation inherited from Turtle, e.g., \texttt{?x a :Software, :Data.} is an abbreviation of two triple patterns \texttt{?x a :Software. ?x a :Data.}.
\item Minimal restrictions on variable and blank nodes names.
\item Prefix notation to abbreviate full URIs.
\item No imposed order on the elements in a graph patterns due to the declarative nature of SPARQL.
\end{itemize}
Thus, it is not possible to analyze recurring patterns in the queries of the dataset using a simple string comparison.
Instead, we propose to analyze recurring patterns using the following signature extraction procedure.

A signature of a SPARQL-OWL query is obtained from the query using the following steps:
\begin{enumerate}
\item Parse the query to a SPARQL algebra expression, expanding all the abbreviations in the process.
\item Remove all the solution modifiers, corresponding to the used verb and projection.
The rationale for this is that this is related more to the form of the original question rather than to its content and can be relatively easily changed, e.g., a question \emph{Is Weka a free software?} can be changed to \emph{What software is a free software?}.
\item From all BGPs in the algebra expression remove triple patterns of form $(\cdot, \texttt{rdf:type}, \texttt{owl:Restriction})$ and $(\cdot, \texttt{rdf:type}, \texttt{owl:Class})$, where $\cdot$ stands for any node.
Triples of such forms are an artifact of OWL serialization to RDF and while they are crucial in the actual ontology file, they are, barring a user actively trying to hack the restricted OWL vocabulary, redundant in the query.
\item From all filters remove expressions of forms \texttt{?var != owl:Nothing}, where \texttt{?var} stands for a variable or a blank node.
If the removed expression was a part of a larger expression (e.g., an operand in conjunction) simplify it.
If this was a single expression in a filter, remove the filter.
The rationale for this is that filtering for unsatisfiable concepts is rather a decision which must be applied consistently to all or to no queries.
\item Remove \texttt{*} and \texttt{+} from all property paths where these symbols refer to  a known transitive property, e.g., \texttt{rdfs:subClassOf} or \texttt{rdfs:subPropertyOf}.
The rationale is this is redundant with entailment regime of SPARQL-OWL.
\item Replace a property paths of form \texttt{rdf:type/rdfs:subClassOf} by \texttt{rdf:type} to remove redundancy.
\item Merge BGPs that are siblings in the algebra expression into a single BGP.
\item Replace all the URIs from namespaces other than RDF, RDFS, OWL and XSD with new blank nodes in a consistent manner, i.e., within a single query the same URI is always replaced by the same blank node.
This decouples the query from the concrete question and allows for generalization.
\end{enumerate}

The described procedure does not take into account variablity in naming variables and blank nodes nor ordering flexibility, and thus causes a query with triple patterns shuffled to have different signature than the original query.
To address this, we introduce the notion of signature equivalence.
Two signatures $S_1$, $S_2$ are equivalent if, and only if, it is possible to find a one-to-one mapping $\sigma$ (i.e., a bijection) from the set of variables and blank nodes in $S_1$ to the set of variables and blank nodes in $S_2$ such that by applying $\sigma$ to $S_2$ one obtains an expression that is isomorphic with $S_1$, i.e., identical to $S_1$ barring the order of operators.

As examples, consider the following two queries.
The first one is CQ6 from AWO, the other is CQ9 from Dem@ware.
For clarity, we do not introduce additional syntax for SPARQL algebra and instead we keep using the SPARQL-OWL syntax, but using complete triple patterns instead of abbreviated ones.

\begin{alltt}
SELECT DISTINCT ?eats
WHERE \{
   ?eats rdfs:subClassOf awo:plant, [
        a owl:Restriction ;
        owl:onProperty awo:eats;
        owl:someValuesFrom awo:animal
   ] .
   FILTER(?eats != owl:Nothing)
\}
\end{alltt}

\begin{alltt}
SELECT DISTINCT *
WHERE \{
    _:c1 rdfs:subClassOf 
      lab:CognitiveAbilitiesAssessment, [
        owl:onProperty ?p;
        owl:someValuesFrom _:c2
    ].
\}
\end{alltt}

Computing the signatures, we obtain, respectively:

\begin{alltt}
?eats rdfs:subClassOf _:b2.
?eats rdfs:subClassOf _:b1.
_:b1 owl:onProperty _:b3.
_:b1 owl:someValuesFrom _:b4.
\end{alltt}

\begin{alltt}
_:c1 rdfs:subClassOf _:c4.
_:c1 rdfs:subClassOf _:c3.
_:c3 owl:onProperty ?p.
_:c3 owl:someValuesFrom _:c2.
\end{alltt}
We observe that there exists a bijection between the sets of variables and blank nodes: $\{\texttt{?eats}\mapsfrom \texttt{\_:c1}, \texttt{\_:b1}\mapsfrom \texttt{\_:c3}, \texttt{\_:b2}\mapsfrom \texttt{\_:c4}, \texttt{\_:b3}\mapsfrom \texttt{?p}, \texttt{\_:b4}\mapsfrom \texttt{\_:c2}\}$, and thus we deem that both signatures are equivalent and the queries can be considered to share a recurring pattern. 

In Table \ref{tab:signatures} we present signatures that are shared by at least three queries in the dataset, sorted by the total number of queries having that signature.
The most common signature was shared between 26 queries originating from 4 ontologies.
The 10 most common signatures presented in Table \ref{tab:signatures} is shared by 86 queries, i.e., $65.6\%$ of all the queries in the dataset (c.f. Table~\ref{table:cq_sparql_translatability}).

To obtain a similar statistics on the ontology level, we can sum the presented numbers for each ontology and compare them with number of queries for the respective ontology.
In this sense, the most diverse is the Stuff ontology, with 0 queries having one of the 9 most common signatures and the Software ontology is next in line with 13 queries (i.e., $33\%$ of all the queries).
AWO has 4 ($57\%$) queries having one of these signatures, while OntoDT has 8 ($62\%$).
Finally, the least diverse is Dem@Care with 57 out of 60 queries ($95\%$) with one of these signatures.
We note that some of the signatures are distinct for Dem@Care and the number obtained for it must be treated with caution.

\begin{table*}
\caption{The signatures that are common for at least three queries in the dataset.
In the column Ontologies listed are ontologies from which the queries originated along with the number of queries having that signature.
}
\label{tab:signatures}
\centering
\begin{tabular}{p{.70\textwidth}p{.22\textwidth}}
\hline
Signature & Ontologies \\
\hline\hline

\texttt{[] rdfs:subClassOf [], [owl:onProperty []; owl:someValuesFrom []]}
&
Dem@Care~(23), AWO~(1), OntoDT~(2), SWO~(1) \\

\hline

\texttt{?x  rdfs:subClassOf  ?y   FILTER ( ?x != ?y )}
&
Dem@Care (17) \\

\hline

\texttt{[] rdfs:subClassOf [], [owl:onProperty  []; owl:someValuesFrom  ?x]. ?x rdfs:subClassOf  [].}
&
SWO (3), OntoDT (6)    \\

\hline

\texttt{[] rdfs:subClassOf  [owl:onProperty  []; owl:someValuesFrom  ?x].\newline
?x rdfs:subClassOf  ?y FILTER ( ?x != ?y )
}
&
SWO(1), Dem@Care (7) \\

\hline

\texttt{?x  rdfs:subClassOf  ?y
	FILTER NOT EXISTS \{ ?x  rdfs:subClassOf  ?z .
		?z rdfs:subClassOf  ?y
		FILTER(?z != ?y \&\& ?x != ?z)
	\}
	FILTER(?x != ?y)
}
&
Dem@Care (7) \\

\hline

\texttt{[] rdfs:subClassOf [ owl:onProperty []; owl:hasValue [] ]}
&
SWO (5) \\

\hline

\texttt{[] rdfs:subClassOf  [], [
	 owl:onProperty [] ;
	owl:hasValue  [] ]
}
&
SWO (4) \\

\hline

\texttt{[] rdfs:subClassOf [owl:onProperty []; owl:somevaluesFrom []] }
&
Dem@Care (1), SWO (2)
\\

\hline

\texttt{[]  a []}
&
Dem@Care (3) \\

\hline

\texttt{[] rdfs:subClassOf ?x, [owl:onProperty  []; owl:someValuesFrom ?y].
?y rdfs:subClassOf  ?x
}
&
AWO (3) \\
\hline
\end{tabular}
\end{table*}

Another approach would be to consider only signatures that are shared between ontologies.
In Table \ref{tab:signatures-two-onts} we present signatures that are shared by queries obtained from at least two ontologies.
There are 6 signatures shared by queries originating from different ontologies and they are shared by 49 queries, i.e., $37.4\%$ of all the queries in the dataset.



We can assess dissimilarity of the obtained queries by computing, for each ontology, how many of the queries  do not share a signature with a query from any other ontology.
All the relevant signatures are gathered in Table \ref{tab:signatures-two-onts}, and using the numbers from Table~\ref{table:cq_sparql_translatability}, we obtain the following:
 SWO, Stuff and AWO perform similarly with, respectively, 34 ($85\%$), 7 ($78\%$) and 5 ($71\%$) queries.
The remaining ontologies have much fewer dissimilar queries, with 29 ($48\%$) in Dem@Care and 5 ($38\%$) in OntoDT.

It is of no surprise that these statistics are not consistent with the similar statistics computed on the linguistic level.
This is yet another reflection of the gap between the questions and queries and of the fact that the form of a query depends on the question as well as on the ontology.

\begin{table*}
\caption{The signatures that are shared between queries coming from at least two ontologies.
In the column Ontologies listed are ontologies from which the queries originated along with the number of queries having that signature.
}
\label{tab:signatures-two-onts}
\centering
\begin{tabular}{p{.70\textwidth}p{.22\textwidth}}
\hline
Signature & Ontologies \\
\hline\hline

\texttt{[] rdfs:subClassOf [], [owl:onProperty []; owl:someValuesFrom []]}
&
Dem@Care~(23), AWO~(1), OntoDT~(2), SWO~(1) \\

\hline

\texttt{[] rdfs:subClassOf [], [owl:onProperty  []; owl:someValuesFrom  ?x]. ?x rdfs:subClassOf  [].}
&
SWO (3), OntoDT (6)    \\

\hline

\texttt{[] rdfs:subClassOf  [owl:onProperty  []; owl:someValuesFrom  ?x].\newline
	?x rdfs:subClassOf  ?y FILTER ( ?x != ?y )
}
&
SWO (1), Dem@Care (7) \\

\hline

\texttt{[] rdfs:subClassOf  [] ;
rdfs:subClassOf [
owl:onProperty  [] ;
owl:someValuesFrom  [
 owl:onProperty [] ;
owl:someValuesFrom  ?x ]].
?x  rdfs:subClassOf ?y
FILTER ( ?x != ?y )
}
&
SWO (1), Stuff (1) \\

\hline

\texttt{[] rdfs:subClassOf [
owl:onProperty [] ;
owl:someValuesFrom  []]
}
&
Dem@Care (1), SWO(1) \\

\hline

\texttt{[] rdfs:subClassOf  [], []}
&
AWO (1), Stuff (1) \\

\hline
\end{tabular}
\end{table*}

\subsection{Mapping between CQ patterns and SPARQL-OWL signatures}
\label{mn}
Our analysis showed that mapping between patterns extracted from CQs and those extracted from collected SPARQL-OWL queries is a many-to-many relation. The single competency question pattern can have multiple SPARQL-OWL signatures and single SPARQL-OWL signature also can have multiple question patterns.

The following reasons justify that observation:

\subsubsection{Multiple sparql-owl signatures for one CQ pattern}
Property restrictions: An ontology engineer can model the ontology using domain and range property restrictions. They specify on which side of the property given class should be placed. Because there is no general rule how to do that, different people can model knowledge in a different way. Thus, even if one CQ pattern is identified in multiple ontologies, we cannot be sure what the translated SPARQL-OWL query will be until the ontology is analyzed.

For example, CQ: \emph{What is the input of Droid?} with vocabulary in ontology: \textsf{input} (property), \textsf{Droid} (named class) can be mapped into:
\begin{enumerate}

\item \texttt{SELECT ?x WHERE \{:Droid :input ?x\} }
\item \texttt{SELECT ?x WHERE \{?x :input :Droid\} }
\end{enumerate}
depending on :input property restrictions. If \textsf{Software} (of which \textsf{Droid} is a subtype) is stated in \texttt{rdfs:domain} of property restriction, query 1 should be produced. If the property mentions a Software in \texttt{rdfs:range}, the second one should be proposed.

Type of property: each property can be of either of data or object type. Depending on the vocabulary in the ontology, different queries may be produced. For example, for CQ: \emph{What is the homepage of Windows?}, if there is a \textsf{URL} class and \textsf{has\_homepage} object property expects a \textsf{URL} in its range, query like:
\begin{alltt}
SELECT ?x WHERE \{
    :Windows rdfs:subClassOf [
        a owl:Restriction ; 
        owl:onProperty :has\_homepage ;
        owl:someValuesFrom ?x
    ]
\} 
\end{alltt} should be constructed. If \textsf{has\_homepage} is a datatype property, query similar to: 
\begin{alltt}
	SELECT ?x WHERE \{
	    :Windows rdfs:subClassOf [
	        a owl:Restriction ;
	        owl:onProperty :has\_homepage ;
	        owl:hasValue ?x
	     ]
	\} 
\end{alltt} should be proposed.

Indirect vocabulary: although there is no matching vocabulary that can be used as translations of chunks and can be used to construct the SPARQL-OWL query, sometimes it is possible to construct a desired concept out of multiple available classes. A real world example comes from the Software Ontology:
For CQ: \emph{What is the alternative to Weka?}
We have a SPARQL-OWL query provided in the dataset for that CQ, but, because there is no good candidate for \textsf{alternative} in an ontology, the engineer who proposed the SPARQL-OWL query used more abstract reasoning. In domain of Software, an alternative can be defined as another software implementing the same algorithms as given one. The SWO ontology against which the CQ was stated has the object property \textsf{implements} as well as \textsf{Weka}', \textsf{Algorithm} and \textsf{Software} classes, so they were used to construct a query instead.


\subsubsection{Multiple CQ patterns for single sparql-owl signature}
Expressivity of natural language - Many questions with the same meaning can be expressed in multiple forms, for instance for SPARQL-OWL signature:  

\begin{lstlisting}
?x rdfs:subClassOf _:b2, [
         owl:onProperty _:b3 ;
         owl:someValuesFrom ?w ] .
?y rdfs:subClassOf _:b2, [
         owl:onProperty  _:b3 ;
         owl:someValuesFrom  ?w ] .
?w  rdfs:subClassOf  ?z
FILTER ( ?w != ?z && ?x != ?y)
\end{lstlisting}

the following CQ patterns are found:
\begin{itemize}
\item What EC1 to EC2 are there?
\item What are EC1 to EC2?
\end{itemize}

\subsection{Signal words and phrases}

The dataset we collected allowed to verify if there is a correspondence between particular words, phrases or sentence structures and vocabulary from SPARQL-OWL queries. In other words, we would like to check if presence of some words or sequences of words signal that some SPARQL-OWL vocabulary or whole query structure should be used. Altough section \ref{mn} showed that even for single CQ pattern multiple SPARQL-OWL queries can be constructed when dealing with differently modelled ontologies, and similarly for single SPARQL-OWL pattern many different CQs can be formulated, we would like to check if some correspondences are so common, that they can help engineers to construct SPARQL-OWL queries and competency questions as recommendation.

In order to verify it, we used the following procedure to extract correspondence information:

\begin{itemize}
\item Group CQs by common words/phrases/sentence structures.
\item Inside each created group, create subgroups of CQs for which the same (ignoring URIs) SPARQL-OWL query was constructed. If given CQ doesn't have tranlation because of lacking vocabulary in an ontology - the CQ is ignored and not analyzed in further steps.
\item If inside of given group, there is a subgroup of size bigger than 1, decide whether link between group and subgroup is meaningful. For instance, if the group is created because multiple CQs share word "the" - the subgroups of shared SPARQL-OWL queries will be accidental.
\item List all interesting cases.
\end{itemize}

In our dataset, we can observe multiple questions asking for subclasses or superclasses of given one. They share similar structure , differing only in ECs used and they share same SPARQL-OWL queries (ignoring URIs).

Table \ref{tab:correspondence} summarizes observed correspondences. Column \textsf{signal} indicate the frequent sequence of words that from our dataset. Wherever synonymical words can be observed in the dataset, they are separated using single slash. Column  \textsf{Corresponding SPARQL-OWL} presents SPARQL-OWL query or fragment that was used extensively for given group. The third column provides information on how many times (out of all provided SPARQL-OWL queries proposed for all CQs from a given group) the SPARQL-OWL pattern from column 2 was used. In all queries, the \texttt{:URI} means some -- possibly different for each usage -- URI.

\begin{table*}[h]
\begin{center}
\caption{Frequent signal phrase with most frequently SPARQL-OWL queries cooccurrences count}
\label{tab:correspondence}.

 \begin{tabular}{p{.28\textwidth}p{.54\textwidth}p{0.1\textwidth}}
 \hline
 Signal & Corresponding SPARQL-OWL & Cooccurrences \\
 \hline\hline
What are the possible types \ldots & \texttt{SELECT DISTINCT * WHERE \newline \{ ?x~rdfs:subClassOf :URI . FILTER(?x != :URI \&\& ?x != owl:Nothing) \} } & 3/3 (100\%) \\
\hline

What are the types of \ldots & \texttt{SELECT DISTINCT * WHERE \newline \{ ?x~rdfs:subClassOf :URI . FILTER(?x != :URI \&\& ?x != owl:Nothing) \} } & 3/4 (75\%) \\

 What types of \ldots is/are \ldots & \texttt{SELECT DISTINCT * WHERE \newline \{ [] rdfs:subClassOf :URI, [ owl:onProperty ?x; owl:someValuesFrom [] ]. \}}  & 8/11 (72.3\%)\\
  \hline

Which/what kind of \dots is/are \ldots &  \texttt{SELECT DISTINCT * WHERE \{ :URI rdfs:subClassOf ?x . ?x~rdfs:subClassOf :URI. FILTER(?x != :URI \&\& ?x != :URI) \} } & 2/3 (66.7\%) \\
 \hline

 What are the main types of \ldots & \texttt{SELECT DISTINCT * WHERE \{ ?x rdfs:subClassOf :URI. FILTER NOT EXISTS \{ ?x rdfs:subClassOf ?y . \newline ?y rdfs:subClassOf :URI. \} \newline FILTER(?x != :URI \&\& ?x != owl:Nothing) \}  } & 6/9 (66\%) \\
 \hline

 \hline

\end{tabular}
\end{center}
\end{table*}

Interestngly, there is no strong connection between signal words like \textsf{or} or \textsf{and} with unions and intersections expressed using SPARQL-OWL. Table \ref{table:correspondence_words} summarizes some interesting cases.  The reason for that is the fact that both words are widely used in different contexts.

We also placed \textsf{exactly NUMBER EC} in the table, even though there is only one example observed, but we strongly believe that such pattern would be shared among multiple cases. In the dataset cardinality restriction was used along with \texttt{owl:cardinality} in SPARQL-OWL translation.
\begin{table*}[h]
\begin{center}
\caption{Signal words}
\label{table:correspondence_words}.

 \begin{tabular}{lp{.35\textwidth}r}
 \hline
 Signal & Corresponding SPARQL-OWL & Support \\
 \hline\hline
 \emph{Which/What/Who/Where/When} -- at the beginning of CQ & SELECT type query & 107/107 (100\%) \\
 \hline
 \emph{Is/Are/Can/Does} -- at the beginning of CQ &  ASK type query & 16/18 (88.9\%) \\
 \hline
 \emph{or} -- used as part of CQ & \texttt{owl:unionOf} -- present in SPARQL-OWL & 2/9 (22.2\%) \\
 \hline
 \emph{and} -- used as part of CQ & \texttt{owl:intersectionOf} -- present in SPARQL-OWL & 2/11 (18.2\%) \\
 \hline
 \hline
 \emph{exactly NUMBER ENTITY} &  \texttt{owl:cardinality "NUMBER"\typedliteral{}xsd:nonNegativeInteger} & 1/1 (100\%) \\
 \hline

\end{tabular}
\end{center}
\end{table*}



\section{Discussion}
\label{sec:disc}

\paragraph{Answering the research questions}
We now return to the research questions stated in the Introduction.

Regarding RQ1, {\em Increasing the scope in domains and ontologies, are there more CQ patterns than those identified in the state-of-art papers?}, this can be answered in the affirmative. More precisely, there are 106 distinct CQ patterns, which are the linguistic patterns of CQs rather than a merger of CQ template based on one modelling style, and still 82 after further normalisation (such as all phrases in the singular). Thus, increasing the size of the CQ data set (cf. related works) with more subject domains and more CQ authors does affect pattern identification.

The linguistic patterns themselves are \textit{not} specific to particular constructs of OWL, therewith answering RQ2 in the negative. Mainly, a PC may well be mapped to an object  property, data property, or a class. 

There are recurring patterns at SPARQL-OWL level, hence, RQ3 can be answered with `yes'. In particular, there are 46 patterns, some of which are query patterns that can be applied to more than one ontology, and the 9 most common query patterns cover 63.1\% of all the queries in the data set. 

With respect to RQ4, we have found that the mapping between lingustic patterns and those extracted from the collected SPARQL-OWL queries is a many-to-many relation. 
More specifically, the linking of the linguistic patterns of CQs to SPARQL-OWL patterns are such that there can be multiple SPARQL-OWL queries for one distinct CQ pattern, which is due principally to the different ways one can represent knowledge in the ontology, and there can be multiple CQ patterns for a single SPARQL-OWL query, as there are different ways to formulate the same thing in natural language. The latter easily could have been expected upfront, for this is a well-known aspect of controlled natural languages and the notion of variation in natural language generation within the scope of natural language interfaces to databases. 

\paragraph{Considerations for future work}

Altough we provided a broad analysis of the collected corpus, there are still some open questions that at the current state of the corpora cannot be answered. The most interesting area for further research may be detecting when it is possible to provide a translation for a CQ when ontology is provided. Our basic analysis resulted in an identification of three cases where proposing such translation is possible:

\begin{itemize}

\item Extracted EC and PCs are available in an ontology, they can be identified as part of the URI or \texttt{rdfs:label}. For example, CQ: stuff\_01 \emph{Is [this stuff] a pure or a mixed stuff?}

with SPARQL-OWL query: 

\begin{alltt}

SELECT DISTINCT *
WHERE \{
  $PPx1$ rdfs:subClassOf ?class.
  FILTER(?class IN (:PureStuff, :MixedStuff))
\}
\end{alltt}
uses only classes that are identified using the same words like in extracted ECs.

\item There is no vocabulary for EC and/or PC in an ontology, but the synonymes can be found and used instead. For example, For example CQ swo14: \emph{Which software tool created [this data]?} with SPARQL-OWL translation provided:

\begin{alltt}
SELECT ?sw WHERE \{ 
  ?sw rdfs:subClassOf swo:SWO_0000001 ;
      rdfs:subClassOf [
        a owl:Restriction ;
        owl:onProperty swo:SWO_0000087 ;   
        owl:someValuesFrom $PPx1$
    ].
    $PPx1$ rdfs:subClassOf obo:IAO_0000027 .
\}
\end{alltt}

There is no class nor property named \emph{created} in ontology but instead, one can use
\texttt{SWO\_0000087} labelled as \emph{has specified data output} to produce query returning expected results.

\item There is no vocabulary for EC and/or RC in an ontology, but we can construct a translation out of multiple entities from the ontology.

For example CQ swo02: \emph{What are the alternatives to [this software]?} translates to:

\begin{alltt}
SELECT ?sw2
WHERE \{
  $PPx1$ rdfs:subClassOf swo:SWO_0000001 , [
     a owl:Restriction ;
     owl:onProperty efo-swo:SWO_0000740 ;
     owl:someValuesFrom ?alg
  ] .
  ?sw2 rdfs:subClassOf swo:SWO_0000001 , [
     a owl:Restriction ;
     owl:onProperty efo-swo:SWO_0000740 ;
     owl:someValuesFrom ?alg
  ] .
  ?alg rdfs:subClassOf obo:IAO_0000064 .
  FILTER(?alg != obo:IAO_0000064 &&
         ?sw1 != ?sw2)
\}
\end{alltt}

which can be interpreted as "What other software implementing the same algorithm are there?" As the synonym of alternative to software is the software with the same functionality.

\end{itemize}

We think that first two situations can be successfully addressed by automated methods, but situation number 3 requires some level of domain understanding to decide whether two things described differently are equal in meaning in the context of ontology domain. 

We encourage ontology engineers to utilize those patterns defined in Tables \ref{tab:shared_cq_patterns} and \ref{tab:shared_cq_patterns_lv3} as good practices worth considering while creating own CQs. Incorporating such good practices may help to produce automatic solutions verifying ontologies without manual translation of CQs into SPARQL-OWL form.

Moreover, we noticed that there are cases in which providing a simple yes/no answer for binary questions may be not enough. Although the yes/no answer for CQ swo05: \emph{Are there any modification to [the algorithm] [the tool] uses?} is correct, it is of limited usefullness, because one may expect the modifications to be listed. Thus instead of ASK query - SELECT may be considered.

\section{Conclusions}
\label{sec:concl}

This paper presented the, to date, largest set of Competency Questions (CQs), consisting of 234 CQ related to 5 ontologies and originating from a range of domains and authored by diverse groups, extending the smaller CQ datasets of prior works. The CQs also have their formalisations in the form of SPARQL-OWL queries for the specific ontology the CQs relate to, where possible, resulting in 131 queries.
 
The analysis of this comprehensive set of CQs has shown that the analysis of CQs in order to find CQ archetypes, cannot be executed as dependent on the actual ontology, which had been done in previous approaches, since CQs are a part of the requirements specification phase in ontology engineering methodologies, and thus they are upfront of actual implementation of the relevant ontology content. To remedy this, our novel analysis is lexico-syntactic, i.e., at the level of natural language rather than assuming pre-conceived modelling patterns in the requirements analysis stage. We have found 106 principal linguistic patterns in the data set, which may be reduced to 82, which is a 4-5-fold increase in patterns previously observed through manual analyses. We have also found 46 recurring patterns at the SPARQL-OWL level. Moreover, this analysis confirmed the hypothesis that there is an $m$:$n$ relation between CQs and their formalisations, due to different ontology modelling styles.

We hope that the dataset may be of use for further research especially into CQs, so that it may be included effectively in ontology development processes. The CQ patterns can be viewed as a prelude to a user and usage-driven CNL for CQs. The SPARQL-OWL queries and their patterns may inform optimisation of their execution or query design interfaces and similar common research activities.




\subsection*{Acknowledgments} 
This work was partly supported by the Polish National Science Center (Grant No 2014/13/D/ST6/02076).
Jedrzej Potoniec acknowledges support from the grant 09/91/DSPB/0627.

\appendix
\section{List of distinct CQ patterns}
\begin{footnotesize}
\noindent What is EC1 PC1 EC2 (*) \\
What are EC1 to EC2\\
What EC1 to EC2 are there\\
Which of the EC1 and EC2 PC1 EC3 PC1\\
Are there any EC1 to EC2 EC3 PC1\\
PC1 EC1 PC1 EC2 (*) \\
What type of EC1 is EC2 (*) \\
What EC1 PC1 EC2 (*) \\
Is EC1 EC2 for EC3\\
What are EC1 and EC2 of EC3\\
Which EC1 is there for EC2 and what PC1 EC3 PC1\\
Which EC1 PC1 EC2\\
What EC1 PC1 I PC1 EC2 PC1 EC3 (*) \\
What are EC1 and EC2 for EC3\\
What EC1 from EC2 PC1 EC3, EC4\\
What are EC1 for EC2\\
What is EC1 for EC2\\
PC1 EC1 PC1 EC2 to EC3 (*) \\
PC1 EC1 PC1 EC2 that are EC3 from EC4\\
To what extent PC1 EC1 PC1 EC2 (*) \\
What EC1 PC1 I PC1 EC2 in EC3\\
Is EC1 of EC2 EC3\\
PC1 I PC1 EC1 if EC2 PC2 EC3 (*) \\
Given EC1, what are EC2 for EC3 of EC4\\
Where PC1 I PC1 EC1 (*) \\
Is there EC1 for EC2\\
How PC1 I PC1 EC1 (*) \\
How PC1 I PC1 EC1 with EC2 PC1\\
Are there any EC1 PC1 EC2 PC1\\
Where PC1 I PC1 EC1 for EC2\\
Who PC1 EC1 (*) \\
What is EC1 of EC2\\
Can we PC1 EC1 of EC2\\
Where PC1 I PC1 EC1 PC1 (*) \\
Which EC1 PC1 I PC1 EC2 PC1\\
Which is EC1 PC1 EC2\\
Do I know EC1 who PC1 EC2 or PC1 EC3 (*) \\
How and where PC1 EC1 PC1 in the past (*) \\
How long PC1 EC1 PC1 (*) \\
How EC1 is EC2 (*) \\
What do EC1 PC1 EC2 EC3\\
What EC1 PC1 EC2 given EC3 (*) \\
Who are EC1 of EC2\\
Who else PC1 EC1 EC2 (*) \\
How many EC1 PC1 I PC1 EC2 (*) \\
PC1 EC1 PC2 EC2 (*) \\
What EC1 are in EC2 of EC3\\
What are the differences between EC1 of EC2\\
When PC1 EC1 of EC2 PC1\\
Is EC1 EC2 (*) \\
What EC1 does EC2 have, and what is its EC3\\
Is EC1 EC2 or not (*) \\
At what EC1 PC1 EC2 of EC3 PC1\\
Who PC1 EC1 for EC2\\
How many EC1 PC1 we PC1 EC2 EC3\\
PC1 I PC1 EC1 PC1 EC2 (*) \\
Does EC1 of EC2 RC1 EC3\\
Are there any EC1 for EC2\\
Is there any EC1 for EC2 and where PC1 I PC1 EC3\\
Does EC1 have EC2\\
Where is EC1 of EC2\\
Where's EC1 of EC2\\
How EC1 PC1 is EC2 for EC3\\
How PC1 I PC1 EC2 (*) \\
Is there EC1 with EC2\\
How PC1 I PC1 EC1 PC1 EC2 (*) \\
PC1 I PC1 some EC1 of EC2 for EC3\\
What EC1 PC1 I PC1 EC2 (*) \\
What EC1 PC1 EC2 PC1 (*) \\
In what EC1 PC1 EC2 PC2 (*) \\
PC1 I PC1 EC1 on EC2\\
What EC1 PC1 I PC1 EC2 on EC3\\
Is EC1 EC2 or EC3 (*) \\
What is the difference between EC1 and EC2 (*) \\
In which EC1 are EC2 in EC3\\
Which kind of EC1 are EC2\\
What kind of EC1 is EC2\\
Where do I categorise EC1 like EC2 (*) \\
Which EC1 PC1 EC2 PC1\\
Which EC1 are EC2 of EC3\\
Are there EC1 in EC2\\
Which EC1 PC1 I PC1 to PC2 EC2\\
In what kind of EC1 PC1 EC2 PC1\\
Which EC1 are EC2\\
PC1 EC1 and EC2 PC1 EC3 (*) \\
What types of EC1 are EC2\\
What are the main types of EC1\\
What are the types of EC1\\
Which are EC1\\
What PC1 EC1 (*) \\
What PC1 EC1 of EC2\\
What EC1 are of EC2 with respect to EC3\\
What EC1 is of EC2 regarding EC3 (*) \\
What EC1 PC1 EC1 or EC2 that PC2 EC3 (*) \\
What EC1 is of EC2 regarding EC3 and EC4 (*) \\
What are the main categories of EC1\\
What EC1 are EC2\\
What are the main types of EC1 EC2 PC1\\
What types of EC1 PC1 EC1\\
What are the possible types of EC1\\
What is EC1 of EC2 for EC3\\
What is EC1 of EC2 that have EC3\\
What is EC1 of EC2 that have EC3 and EC4\\
What are EC1 that have EC2\\
What is EC1 of EC2 that have EC3 as EC4\\
What is EC1 of EC2 that PC1 EC3\\ 
What is EC1 to EC2 (*) \\
What EC1 to EC2 is there (*) \\
What of the EC1 and EC2 PC1 EC3 PC1 (*) \\
Is there any EC1 to EC2 EC3 PC1 (*) \\
What is EC1 and EC2 (*) \\
What EC1 is there for EC2 and what PC1 EC3 PC1 (*) \\
What is EC1 (*) \\
PC1 EC1 PC1 EC2 that is EC3 from EC4 (*) \\
Given EC1, what is EC2 (*) \\
Is there EC1 (*) \\
How PC1 I PC1 EC1 PC1 (*) \\
Is there any EC1 PC1 EC2 PC1 (*) \\
Can I PC1 EC1 (*) \\
What EC1 PC1 I PC1 EC2 PC1 (*) \\
Which is EC1 (*) \\
What do EC1 PC1 EC2 (*) \\
Who is EC1 (*) \\
What EC1 is in EC2 (*) \\
What is the difference between EC1 (*) \\
When PC1 EC1 PC1 (*) \\
What EC1 do EC2 have, and what is its EC3 (*) \\
At what EC1 PC1 EC2 PC1 (*) \\
How many EC1 PC1 I PC1 EC2 EC3 (*) \\
Do EC1 RC1 EC2 (*) \\
Is there any EC1 (*) \\
Is there any EC1 and where PC1 PC1 EC2 (*) \\
Do EC1 have EC2 (*) \\
Where is EC1 (*) \\
Where's EC1 (*) \\
How EC1 PC1 is EC2 (*) \\
PC1 I PC1 some EC1 (*) \\
PC1 I PC1 EC1 (*) \\
In which EC1 is EC2 (*) \\
What EC1 is EC2 (*) \\
In what type of EC1 PC1 EC2 PC1 (*) \\
What is the main type of EC1 (*) \\
What is the type of EC1 (*) \\
What EC1 PC1 EC2 and EC3 (*) \\
What EC1 is of EC2 with respect to EC3 (*) \\
What is the main type of EC1 EC2 PC1 (*) \\
What type of EC1 PC1 EC1 (*) \\
What is the possible type of EC1 (*) \\
What is EC1 that have EC2 (*) \\
What is EC1 that have EC2 and EC3 (*) \\
What is EC1 that PC1 EC2 (*) \\

\end{footnotesize}



\bibliographystyle{elsarticle-num}

\end{document}